\documentclass{article} 
\usepackage{iclr2027_conference,times}

\usepackage{amsmath,amsfonts,bm}

\def\eqref#1{equation~\ref{#1}}

\def\1{\bm{1}}

\DeclareMathAlphabet{\mathsfit}{\encodingdefault}{\sfdefault}{m}{sl}
\SetMathAlphabet{\mathsfit}{bold}{\encodingdefault}{\sfdefault}{bx}{n}

\usepackage{tabularx}
\usepackage{hyperref}
\usepackage{url}
\usepackage{times}
\usepackage[table]{xcolor}
\usepackage{multicol}
\usepackage{graphicx}
\usepackage{booktabs}
\usepackage{subcaption}
\usepackage{multirow}
\usepackage{float}
\usepackage{adjustbox}
\usepackage{comment}
\title{

OneFixer: High-Quality and Consistent One-Step Autoregressive 3DGS Refinement for Driving Scenes
}

\author{
Boseong Jeon$^{1}$ \quad
Junhyeop Lee$^{2}$ \quad
Juhan Cha$^{1}$ \quad
Hayoung Kim$^{1}$ \\[0.4em]
$^{1}$42dot \qquad
$^{2}$Hyundai Motor Company \\[0.2em]
}
\iclrfinalcopy
\begin{document}

\maketitle

\begin{abstract}
Autoregressive video diffusion is a promising render-time fixer for 3D Gaussian
Splatting (3DGS) in autonomous-driving simulation, but deployment demands high
visual quality and temporal consistency at low latency.
This is particularly challenging for one-step causal generation, where each
imperfect prediction immediately becomes context for subsequent frames.
Existing approaches improve test-time rollouts through staged training with
multiple trained modules and rollout-aware regularization, 
yet one-step
refinement quality can still fall short of what render-time deployment requires.
We introduce \textbf{OneFixer}, a \textbf{one-step} autoregressive
video-diffusion fixer trained in \textbf{one task-specific adaptation stage}.
Our key idea is a \emph{deployment-matched shared rollout}: the model's own
one-step predictions serve as the causal context for flow matching, exposing
training to deployment-time errors, while the same rollout receives direct
pixel-space perceptual supervision to preserve fine detail.
Because the predictions optimized for current-frame quality are exactly those
reused as future context, fidelity and autoregressive robustness are learned
jointly, without bidirectional-to-causal conversion or teacher--student
distillation.
Targeting driving simulation, OneFixer further leverages cues that setting
readily provides---lane geometry and dynamic-agent states---to improve
geometric fidelity.
Across Waymo and proprietary driving scenes with 900-frame autoregressive
rollouts, OneFixer achieves the lowest FVD, LPIPS, and DISTS among all
baselines under one-step inference, with strong geometric fidelity and
temporal consistency matching or exceeding multi-stage DMD pipelines.
Under identical backbone and conditioning, it matches the final quality of
a multi-stage DMD-with-Self-Forcing pipeline in under half the GPU-hours
and continues to improve beyond its plateau.
In closed-loop simulation with a driving policy, OneFixer reduces the
collision rate by a third relative to the raw 3DGS rendering.
Project page: \url{https://onefixer-web.vercel.app/}.
\end{abstract}

\section{Introduction}

3D Gaussian Splatting (3DGS) offers autonomous-driving simulation an explicit,
controllable scene representation with fast rasterized
rendering~\citep{kerbl2023gaussians,zhou2024drivinggaussian}, but novel-view quality degrades severely in driving scenes, where
ego-centric capture along a single trajectory leaves regions sparsely
observed and dynamic objects introduce occlusions and moving
geometry~\citep{zhou2025flexdrive,li2025mtgs,sun2025splatflow}.
Diffusion-based fixers restore realistic appearance from such imperfect
renderings using learned priors~\citep{wu2025difix3d,wu2026realitybridge}, and
among them autoregressive (AR) video
diffusion~\citep{fischerndartifixerenhancing,nvidia2026nvidiaomnidreams,
zhou2026xiaomiauto} is the natural fit for an online render-time fixer:
it is more temporally consistent than image-based
fixers~\citep{wu2025difix3d,zhang2026diffusionharmonizer,wei2025gsfix3d} and,
unlike bidirectional video models~\citep{liu20243dgs}, streams without a
bounded rollout horizon.

\begin{figure*}[t]
    \centering
    \includegraphics[width=0.99\textwidth]{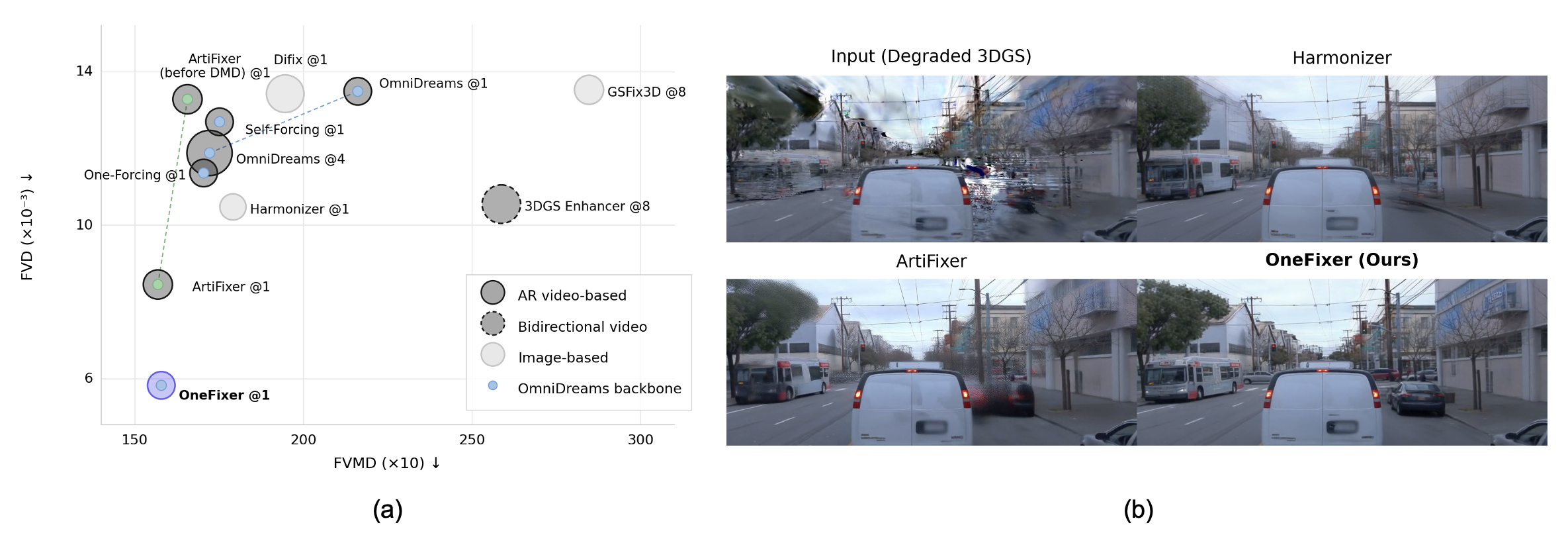}
\caption{
\textbf{Quality--consistency--latency trade-off of 3DGS refinement.}
(a) Methods are positioned by video fidelity (I3D FVD, $\downarrow$) and
temporal consistency (FVMD, $\downarrow$), with bubble area proportional to
inference latency; lower-left and smaller are better.
Dashed lines connect a method across its step-reduction or distillation
path (e.g., OmniDreams @4 $\rightarrow$ @1).
(b) Qualitative comparison on a degraded Waymo 3DGS rendering; OneFixer
recovers finer foliage and a more faithful vehicle under one-step causal
refinement.
}
    \label{fig:tradeoff}
\end{figure*}

Achieving both perceptual quality and temporal consistency in the low-latency,
one- or few-step regime of AR generation is, however, challenging.
Reducing the denoising budget degrades each prediction, and because that
prediction becomes the causal context for the next chunk, errors compound: the
history drifts away from the ground-truth histories seen during training, an
exposure bias well documented in AR video
diffusion~\citep{chen2024diffusionforcing,yin2025slowbidirectional,
huangndselfforcing,li2026rollingsink,feng2026oneforcing}.
Fig.~\ref{fig:tradeoff}(a) illustrates this on causal OmniDreams: moving from
four steps to one substantially lowers latency but worsens both FVD and FVMD,
even though every step is anchored by a 3DGS rendering, which limits drift
only partially.

Prior AR 3DGS fixers address these two failures with separate machinery.
Against drift, they adopt multi-stage training that combines
DMD~\citep{yin2024dmd} with Self-Forcing~\citep{huangndselfforcing}, as in
ArtiFixer and the Xiaomi EV World
Model~\citep{fischerndartifixerenhancing,zhou2026xiaomiauto}; these recipes
are effective but require a bidirectional teacher and carry their own
design burden~\citep{zhao2026causal,you2026adaptivevideodistillation,
luo2026policy}.
Against the quality loss of one-step prediction, they append critics or
reward-based post-training~\citep{feng2026oneforcing,
zhang2026rewardforcing,zhang2026astrolabe,wu2026realitybridge}, each adding
a model or a stage.
Notably, all of these recipes were designed for generic text- or
image-to-video generation from pure noise, and are transferred to 3DGS
refinement without exploiting the structure the task provides.
This raises the central question of this work:
\emph{when the AR task is 3DGS refinement, can we obtain high-quality,
temporally consistent one-step causal generation with a simpler,
single-stage adaptation procedure?}

We introduce OneFixer, which fine-tunes a pretrained causal video
generator~\citep{nvidia2026nvidiaomnidreams} into the deployment-time one-step
3DGS fixer in a single stage, with no
bidirectional-to-causal conversion, teacher--student distillation, critic, or
separate post-training: the generator rolls out its own one-step predictions
exactly as at deployment, and this rollout serves as the causal history
for flow matching and as the source of predictions for direct pixel-space
supervision.
We observe that 3DGS refinement needs no proxy for this rollout.
Because the flow source is the opacity-mixed 3DGS
rendering~\citep{fischerndartifixerenhancing} rather than noise, a single
flow step already yields a plausible, though imperfect, prediction, and because
deployment is also one step, the training rollout
\emph{is} the sampler rather than an approximation of it, unlike multi-step
denoising trajectories or re-noised ground
truth~\citep{huangndselfforcing,guo2026endend}.
Flow matching thus conditions on the model's true deployment-time errors, and
pixel supervision reaches decoded pixels through a single forward pass, with
no truncated backpropagation through a denoising
chain~\citep{wang2026diffusiondrf,prabhudesai2023alignprop}.

Both objectives are optimized \emph{jointly} on this rollout: flow matching
enforces consistency in latent space at every position, and the pixel loss
recovers detail and flows back through the history that produced it, so
preceding predictions are trained as context.
A sequential recipe provides neither coupling~\citep{zhang2026astrolabe,he2026ar}.

OneFixer also exploits cues that driving simulation already exposes: the
opacity-derived reliability of the rendering and, where present, lane geometry
and dynamic-agent states~\citep{yan2024street,li2025mtgs,luo2023latr}.
These matter beyond appearance, since a displaced lane or hallucinated vehicle
is an incorrect observation for the policy under test.

We evaluate OneFixer on Waymo~\citep{sun2020scalability} and proprietary
driving scenes with 900-frame autoregressive rollouts ($9.7\times$ the
training horizon).
OneFixer achieves the lowest FVD, LPIPS, and DISTS at one step with temporal
consistency matching multi-stage DMD pipelines; a same-backbone controlled
study attributes the gain to shared-rollout training, and closed-loop
simulation on 60 scenes shows that the refined observations yield fewer
collisions and off-road events for a driving policy.

\section{Related Work}
\subsection{Diffusion Models for Render-Time 3DGS Enhancement}

Diffusion priors complement explicit 3DGS reconstruction either as
supervision for the scene representation~\citep{wu2025difix3d,liu20243dgs}
or as a render-time fixer that refines rendered observations
directly~\citep{fischerndartifixerenhancing,wu2026realitybridge}; we focus
on the latter.
Image-based fixers such as DiFix3D+~\citep{wu2025difix3d} and
GSFix3D~\citep{wei2025gsfix3d} apply strong image priors to individual
renderings and achieve high frame-level quality, and recent work extends
them with short temporal conditioning for online
use~\citep{zhang2026diffusionharmonizer}; without a video prior, however,
they cannot robustly recognize transient interference such as a Gaussian
floater popping into view and tend to refine it as scene content
(Fig.~\ref{fig:qualitative_comparison}).
Bidirectional video diffusion offers a stronger temporal prior, as in
3DGS-Enhancer~\citep{liu20243dgs}, but operates on fixed clips well short
of minute-scale driving simulation and cannot stream.
Driving-oriented fixers therefore adopt causal autoregressive models:
ArtiFixer~\citep{fischerndartifixerenhancing},
OmniDreams~\citep{nvidia2026nvidiaomnidreams}, and the Xiaomi EV World
Model~\citep{zhou2026xiaomiauto} obtain efficient generators through staged
causal adaptation and distillation, and
RealityBridge~\citep{wu2026realitybridge} follows a supervised curriculum
with reward-guided post-training.
Nevertheless, none establishes quality-preserving one-step deployment,
relying instead on few-step sampling and multi-stage training.

\subsection{Autoregressive Video Diffusion and Exposure Bias}

Causal video generation suffers an exposure bias: training histories come from
clean data while inference conditions on the model's own
predictions~\citep{chen2024diffusionforcing,yin2025slowbidirectional,
huangndselfforcing,li2026rollingsink}.
The prevailing remedy distills a pretrained bidirectional teacher into a causal
few-step student with DMD-style objectives and then exposes the student to its
own rollouts~\citep{yin2025slowbidirectional,huangndselfforcing,
zhu2026causalforcing,zhao2026causal,zheng2026causal}.
This recipe presupposes a bidirectional teacher and is sensitive to student
initialization~\citep{zhu2026causalforcing,zhao2026causal} and distillation
regularization~\citep{you2026adaptivevideodistillation}; even then one-step
quality lags, prompting auxiliary critics or adversarial
post-training~\citep{feng2026oneforcing,wang2026seedvr2}, whose min--max
objective is itself known to be unstable~\citep{wang2026seedvr2}, or
reward-based follow-up stages~\citep{zhang2026rewardforcing,
zhang2026astrolabe,he2026ar}, each adding a model or a stage, and preserving
long-rollout capability through these stages is not
guaranteed~\citep{zhang2026astrolabe,he2026ar}.
Closest to our motivation, self-resampling~\citep{guo2026endend} trains an
autoregressive model without a teacher by generating its own causal context,
but that context is a proxy: ground truth is noised and re-denoised in one
step, so leaked ground-truth content masks the model's true errors and the
training rollout differs from the multi-step sampler used at inference.
It also provides no signal on output pixel quality.
OneFixer instead rolls out the exact one-step deployment generator, so the
history carries the model's true errors, and adds paired pixel-space
supervision on the same rollout.

\subsection{Direct Supervision for One-Step Fidelity}
Recovering perceptual quality at very low step counts typically requires
supervision beyond the diffusion objective.
Image models pioneered this by backpropagating pixel- or reward-space losses
through a truncated denoising chain~\citep{clark2024draft,
prabhudesai2023alignprop} or a single step~\citep{parmar2024onestep,
wu2024osediff,zhang2024s3diff,kim2026fidesr}.
Video models follow the same principle in the bidirectional setting:
DOVE~\citep{chen2025dove} and OASIS~\citep{guo2025oasis} pair pixel-space
supervision with flow-based temporal losses, since frame-wise supervision
alone does not enforce consistency, SeedVR2~\citep{wang2026seedvr2} uses
feature matching and adversarial post-training, and
ChopGrad~\citep{rivkin2026chopgrad} makes latent-space pixel losses tractable
through truncated decoder backpropagation, including for 3DGS refinement.
OASIS additionally starts its single step from the degraded latent rather than
noise, a structured source akin to ours, but in a clip-level model.
Streaming one-step restoration has also been reached, but through multi-stage
distillation from a bidirectional teacher~\citep{zhuang2025flashvsr}.
None of these couples temporal consistency and paired pixel fidelity in a
single stage for autoregressive generation: the bidirectional methods have no
causal history to reuse, while causal pipelines train the reused prediction
through distribution matching or distillation rather than against its clean
reference~\citep{huangndselfforcing,zhuang2025flashvsr}.

\begin{figure*}[t]
    \centering
    \includegraphics[scale=0.95]{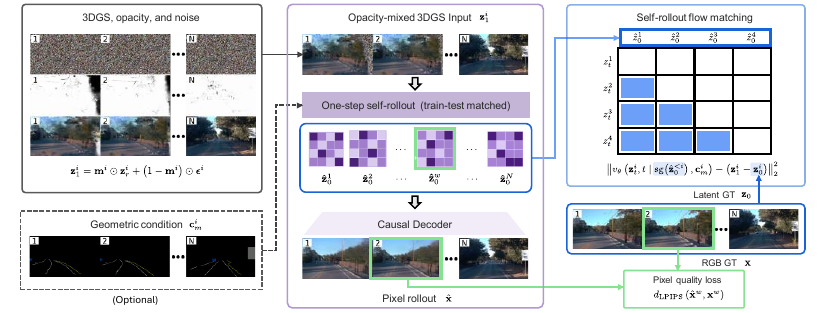}
\caption{
\textbf{Shared-rollout training for one-step 3DGS refinement.}
From opacity-mixed 3DGS inputs and scene conditioning, the deployment-time
one-step causal generator rolls out its own predictions
$\{\hat{\mathbf{z}}_0^1,\ldots,\hat{\mathbf{z}}_0^N\}$ exactly as at
inference.
The rollout serves two objectives: self-rollout flow matching supervises
every position in latent space conditioned on the self-generated history
(detached, $\mathrm{sg}$), and a frame-wise pixel-space perceptual loss on
one sampled prediction back-propagates through the last $k$ latent positions
of that history to recover fine detail.
}
    \label{fig:shared_rollout_training}
\end{figure*}

\section{Method}

\begin{figure*}[t]
    \centering
    \includegraphics[width=0.96\textwidth]{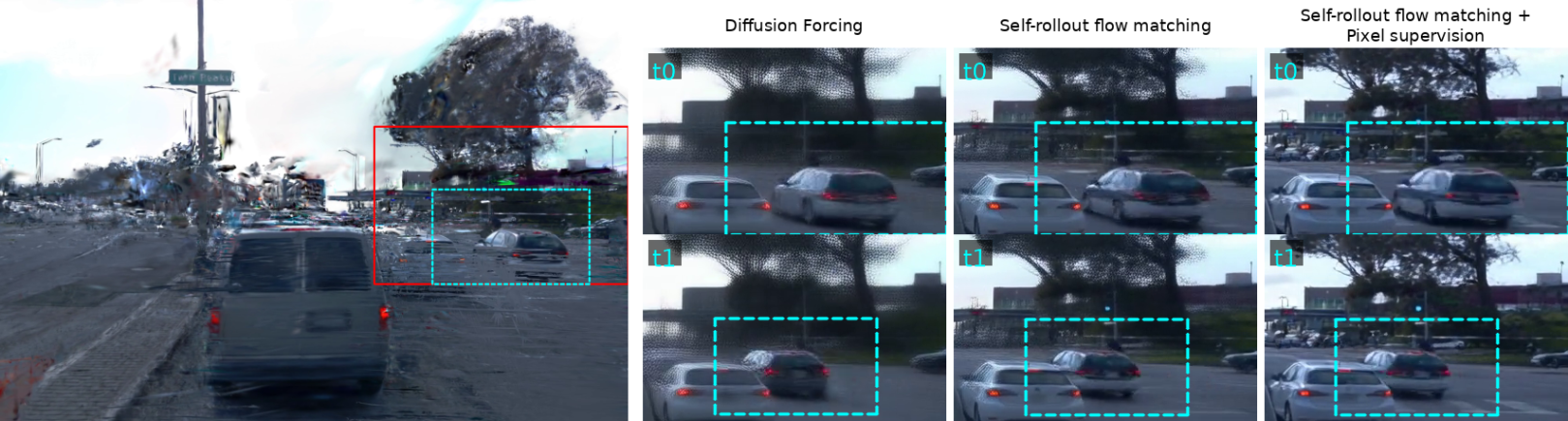}
\caption{
\textbf{Complementary effects of self-rollout and perceptual supervision.}
Left: a Waymo example and its degraded 3DGS input.
Right: one-step autoregressive rollouts and magnified regions.
Diffusion Forcing exhibits appearance variation across generated frames;
self-rollout flow matching improves temporal consistency, while direct
pixel-space perceptual supervision further recovers fine visual details,
particularly in the tree branches and foliage.
}
\label{fig:self_rollout_qualitative}
\end{figure*}

Fig.~\ref{fig:shared_rollout_training} summarizes OneFixer: starting from the
pre-DMD causal OmniDreams checkpoint~\citep{nvidia2026nvidiaomnidreams}, we
train two objectives on one deployment-matched rollout, self-rollout flow
matching (Sec.~\ref{sec:selfrollout}) and frame-wise perceptual supervision
(Sec.~\ref{sec:pixel}); Fig.~\ref{fig:self_rollout_qualitative} previews
their complementary effects.

\subsection{One-Step Causal Refinement from a 3DGS Source}
\label{sec:formulation}
At generation position $i$, the model is conditioned on the causal history
$\mathbf{h}^{<i}$ of preceding latents and a map-derived scene condition
$\mathbf{c}^i_m$ encoding lane geometry and dynamic actors, following
OmniDreams~\citep{nvidia2026nvidiaomnidreams}.
Let $\mathbf{z}^i_r$ and $\mathbf{z}^i_0$ denote the latents of the 3DGS
render and the ground-truth frame, and $\mathbf{m}^i \in [0,1]$ the opacity-derived
reliability mask~\citep{fischerndartifixerenhancing}.
Rather than pure noise $\boldsymbol{\epsilon}^i \sim \mathcal{N}(\mathbf{0},
\mathbf{I})$, the flow source is
\begin{equation}
\mathbf{z}^i_1 = \mathbf{m}^i \odot \mathbf{z}^i_r
  + (1-\mathbf{m}^i) \odot \boldsymbol{\epsilon}^i ,
\label{eq:source}
\end{equation}
which retains rendered content where the reconstruction is reliable and
leaves generative freedom where it is not.
The backbone is a causal flow-matching model: along the interpolant
$\mathbf{z}^i_t = (1-t)\,\mathbf{z}^i_0 + t\,\mathbf{z}^i_1$, a velocity
network $v_\theta(\mathbf{z}^i_t, t \mid \mathbf{h}^{<i}, \mathbf{c}^i_m)$
is trained to predict $\mathbf{z}^i_1 - \mathbf{z}^i_0$, and sampling
integrates from $t{=}1$ to $t{=}0$.
OneFixer performs a single flow step from this source,
\begin{equation}
\hat{\mathbf{z}}^i_0 = G_\theta(\mathbf{z}^i_1 \mid \mathbf{h}^{<i},
  \mathbf{c}^i_m)
  = \mathbf{z}^i_1 - v_\theta(\mathbf{z}^i_1, 1 \mid \mathbf{h}^{<i},
  \mathbf{c}^i_m).
\label{eq:onestep}
\end{equation}
Because the source already carries scene structure, this single step is a
correction rather than a synthesis from noise, which is what makes the
self-rollout of Sec.~\ref{sec:selfrollout} viable; replacing the mixed
source with pure noise and channel-concatenated 3DGS latents, the
alternative explored in ArtiFixer~\citep{fischerndartifixerenhancing},
degrades every metric (V6, Table~\ref{tab:onefixer-waymo-ablation}).

\subsection{Self-Rollout Flow Matching}
\label{sec:selfrollout}

At deployment, $\mathbf{h}^{<i}$ consists of the model's own preceding
one-step predictions $\hat{\mathbf{z}}_0$.
We construct the training history the same way, applying
Eq.~(\ref{eq:onestep}) recursively,
\begin{equation}
\hat{\mathbf{z}}^i_0 = G_\theta\big(\mathbf{z}^i_1 \mid
  \operatorname{sg}(\hat{\mathbf{z}}^{<i}_0), \mathbf{c}^i_m\big),
  \qquad i = 1,\ldots,N,
\label{eq:rollout}
\end{equation}
with $\operatorname{sg}(\cdot)$ the stop-gradient.
For each position we sample $t \sim \mathcal{U}(0,1)$ and minimize
\begin{equation}
\mathcal{L}_{\mathrm{FM}} = \mathbb{E}_{i,t}\Big[\big\|
  v_\theta(\mathbf{z}^i_t, t \mid \operatorname{sg}(\hat{\mathbf{z}}^{<i}_0),
  \mathbf{c}^i_m) - (\mathbf{z}^i_1 - \mathbf{z}^i_0)\big\|_2^2\Big].
\label{eq:fm}
\end{equation}
The flow objective is unchanged; only its causal context is replaced by the
self-generated history, and it supervises every position of the sequence in
latent space, which is where its effect on temporal consistency comes from
(Fig.~\ref{fig:self_rollout_qualitative}, middle).

\subsection{Frame-Wise Pixel Supervision on the Shared Rollout}
\label{sec:pixel}

Latent flow matching supervises the one-step output only in latent space
and never sees the decoded pixels, so it does not directly optimize
perceptual quality.
We therefore sample a single latent position $w \sim \mathcal{U}\{1,\ldots,N\}$
from the training horizon, causally decode $\hat{\mathbf{z}}^w_0$ to RGB, and
minimize
\begin{equation}
\mathcal{L}_{\mathrm{perc}} = d_{\mathrm{LPIPS}}(\hat{\mathbf{x}}^w,
  \mathbf{x}^w),
\label{eq:perc}
\end{equation}
where, with a slight abuse of notation, $\hat{\mathbf{x}}^w$ and
$\mathbf{x}^w$ denote the several RGB frames that the decoder's temporal
compression maps to latent position $w$ and their paired clean references;
the loss is averaged over them.
Because each chunk is a single forward pass, this loss reaches the generator
with no backpropagation through a denoising chain; the only truncation is
across autoregressive history and the causal Wan decoder, whose temporal
cache is detached in the manner of ChopGrad~\citep{rivkin2026chopgrad}.
On the autoregressive side, gradients are retained through the last $k$
latent positions ending at $w$ and earlier history is detached.
The truncation bounds memory (App.~\ref{app:memory}), but the retained path
is also a second channel for consistency: the loss on $\hat{\mathbf{x}}^w$
reaches the $k$ preceding predictions through their role as $w$'s history,
so they are trained to be good \emph{context}, not only good frames.
The ablation confirms that each channel improves FVMD on its own and that
the full model needs both (Sec.~\ref{sec:ablation}).
The full objective is
\begin{equation}
\mathcal{L} = \mathcal{L}_{\mathrm{FM}} + \lambda_{\mathrm{perc}}
  \mathcal{L}_{\mathrm{perc}},
\label{eq:total}
\end{equation}
with both terms evaluated on the same shared rollout.

\newcommand{\best}[1]{\textbf{#1}}
\newcommand{\second}[1]{\underline{#1}}

\section{Results}
\label{sec:results}

\subsection{Experimental Setup}
\label{sec:exp_setup}

\paragraph{Training data and setup.}
We train OneFixer separately on Waymo~\citep{sun2020scalability}
and our proprietary Seoul driving dataset, using 350 and 400
training scenes, respectively.
Following prior 3DGS enhancement
pipelines~\citep{wu2025difix3d,fischerndartifixerenhancing,wu2026realitybridge},
we construct paired degraded-render/ground-truth samples using
cycle-rendered views and under-trained 3DGS reconstructions.
For both domains, we initialize from the official OmniDreams
causal checkpoint released before DMD
distillation~\citep{nvidia2026nvidiaomnidreams}.
Training uses $544 \times 960$ resolution, an effective batch
size of 64, and 93-frame causal sequences.
On Seoul, these sequences are sampled from 300-frame training
source clips.
We use AdamW with a learning rate of $3 \times 10^{-5}$ and
$\lambda_{\mathrm{perc}}=10$, with 400 optimization steps on
Waymo and 600 on Seoul.
Appendix~\ref{app:implementation} provides data-generation and
training details.

\paragraph{Evaluation protocol.}
We evaluate on 80 held-out Waymo scenes using 198-frame
autoregressive rollouts and 32 held-out Seoul scenes using
900-frame rollouts.
These correspond to approximately $2.1\times$ and $9.7\times$
the 93-frame causal training horizon.
We report complementary perceptual, geometric, and temporal
metrics.
Definitions, aggregation rules, and evaluation windows are
provided in Appendix~\ref{app:metrics}.

\paragraph{Baselines and implementation.}
We compare with image-based fixers
DiFix3D+~\citep{wu2025difix3d},
DiffusionHarmonizer~\citep{zhang2026diffusionharmonizer},
and GSFix3D~\citep{wei2025gsfix3d};
the bidirectional video model
3DGS-Enhancer~\citep{liu20243dgs};
and causal video fixers
ArtiFixer~\citep{fischerndartifixerenhancing}
and OmniDreams~\citep{nvidia2026nvidiaomnidreams}.
Each method uses its officially released checkpoint or
recommended pretrained initialization for task-specific adaptation.
For DiffusionHarmonizer we report both its frame-wise mode and a temporal
mode; because the official teacher-forced temporal recipe drifted on our
long rollouts, the temporal variant is trained with self-generated context
(App.~\ref{app:baselines}).
We distinguish three baselines using the OmniDreams backbone.
\emph{OmniDreams} starts from the released DMD-distilled causal
checkpoint and undergoes task-specific diffusion-forcing fine-tuning,
following the fixer adaptation process in the original paper.
\emph{Self-Forcing} follows the staged distillation recipe of
ArtiFixer~\citep{fischerndartifixerenhancing}\footnote{ArtiFixer's project
page describes two phases, but its official implementation trains three
stages: supervised fine-tuning, diffusion-forcing adaptation from the
stage-1 checkpoint, and DMD initialized from the stage-2 student and the
stage-1 critic; we follow the implementation's count.}:
the bidirectional teacher is fine-tuned from the official pre-DMD
bidirectional checkpoint to 3DGS refinement (Stage~1); the causal student
is fine-tuned from the official pre-DMD causal checkpoint---OneFixer's
initialization---to the fixer task (Stage~2); the Stage-2 student is then
distilled with DMD and Self-Forcing~\citep{yin2024dmd,huangndselfforcing}
against the Stage-1 teacher (Stage~3).
ArtiFixer converts its Stage-1 model to causal in Stage~2 because Wan2.1
has no causal release; we instead start Stage~2 from the released causal
checkpoint, as in OmniDreams' own distillation pipeline, so that the
student shares OneFixer's initialization.
\emph{One-Forcing} shares Stages~1--2 and adds the One-Forcing
critic~\citep{feng2026oneforcing} in Stage~3.
Details are in Appendix~\ref{app:baselines}.

\begin{figure*}[t]
    \centering
    \includegraphics[width=0.95\textwidth]{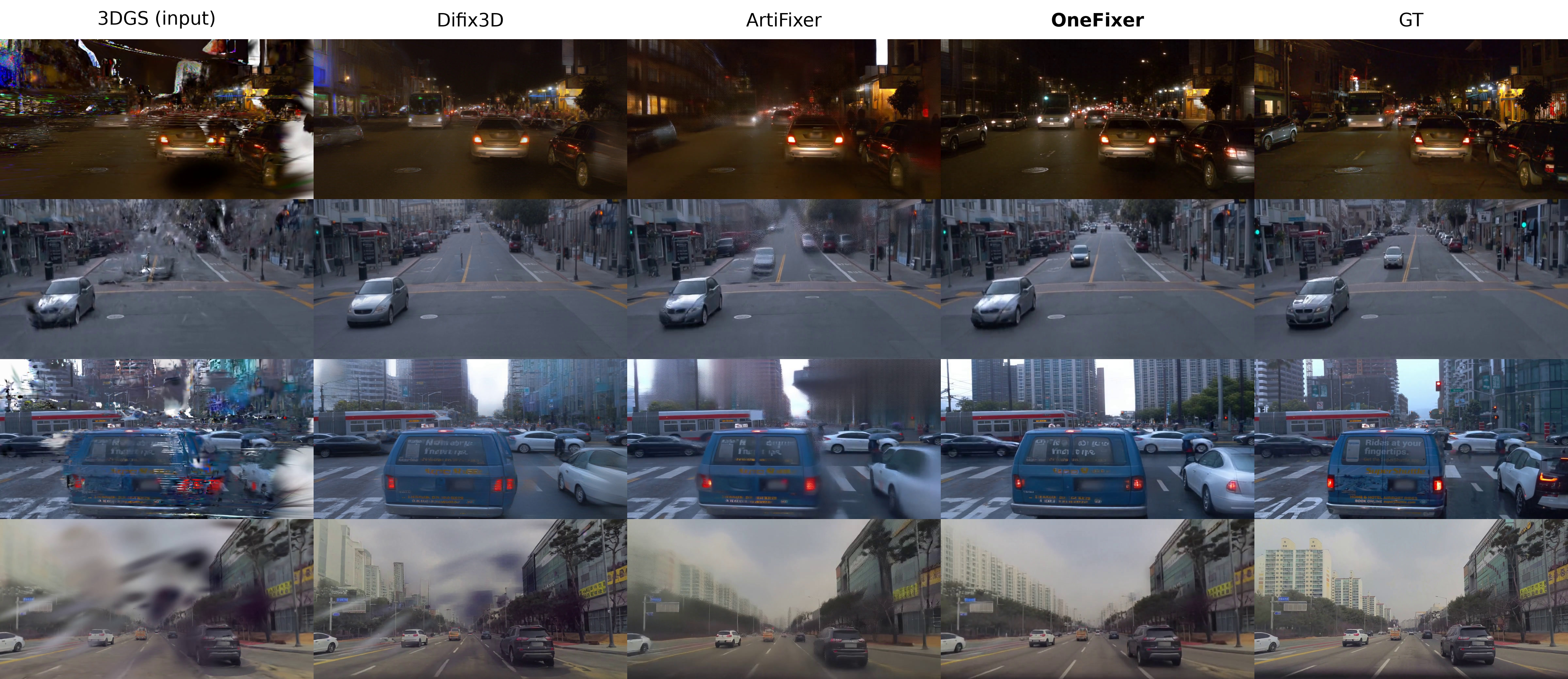}
\caption{
\textbf{Qualitative comparison of 3DGS refinement methods.}
OneFixer better preserves realistic appearance and scene structure under
challenging reconstruction errors.
Its geometry-aware conditioning improves structural fidelity, while the video
prior increases robustness to transient artifacts that are difficult to handle
independently frame by frame, such as the popping Gaussian floater in the
bottom row.
}
    \label{fig:qualitative_comparison}
\end{figure*}

\begin{figure*}[t]
    \centering
    \includegraphics[width=0.96\textwidth]{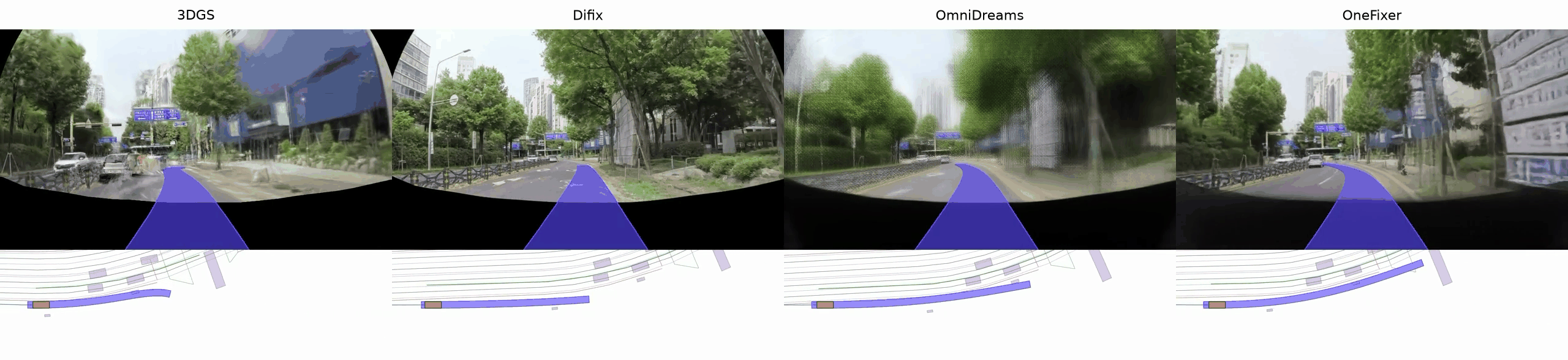}
\caption{
\textbf{Closed-loop replay with different renderers.}
Top: policy observation and planned path (blue) where the 3DGS
reconstruction degrades. Bottom: ego trajectory over the lane map.
Raw 3DGS shows ghosted structure that bends the trajectory; DiFix smears
the lane markings across the road and the policy leaves it; OmniDreams blurs the scene at one step.
OneFixer's faithful observation keeps the policy on route.
}
    \label{fig:closed_loop_qualitative}
\end{figure*}

\begin{figure*}[t]
    \centering
    \includegraphics[width=0.98\textwidth]{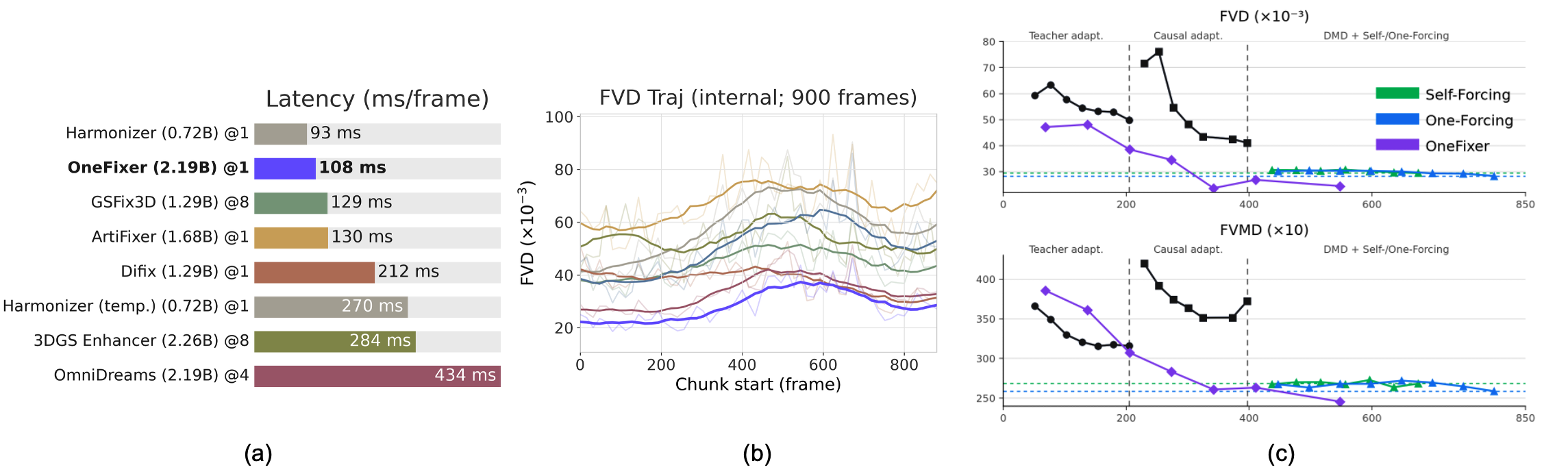}
    \caption{
    \textbf{Inference efficiency, long-horizon stability, and training efficiency.}
    (a) Per-frame enhancement latency measured on a single H200 GPU and bf16 precision for all models.
    @N denotes the number of denoising steps, and model parameter counts
    (in billions) are shown in parentheses.
    (b) Position-wise FVD over 900-frame autoregressive rollouts on the
    proprietary dataset. Self-Forcing is shown in steel blue; other colors match (a).
(c) Validation FVD and FVMD vs cumulative GPU-hours,
32 Waymo scenes, 93-frame one-step rollouts, same backbone and conditioning. Baselines follow the standard fixer-distillation recipe~\citep{fischerndartifixerenhancing, zhou2026xiaomiauto}: teacher adaptation, causal fixer adaptation, then DMD with Self-Forcing (green) or One-Forcing (blue).}
    \label{fig:computation_efficiency}
\end{figure*}

\subsection{Comparison with 3DGS Enhancement Methods}

\paragraph{Overall comparison.}
Table~\ref{tab:fixer-vs-baselines} shows that OneFixer achieves the
strongest one-step balance across perceptual, geometric, and temporal
metrics: it improves temporal consistency over image-based fixers while
retaining their frame-level fidelity, and improves perceptual quality over
bidirectional and causal video fixers while matching or exceeding them on
temporal metrics.
Harmonizer's temporal variant, trained with a warping loss, wins Warp L1 on
both datasets but is otherwise no better than its frame-wise counterpart,
so short-range smoothness alone does not yield video-level consistency.
On Waymo, the same-backbone Self-Forcing and One-Forcing baselines attain
the best PSNR and SSIM but trail OneFixer on FVD, LPIPS, DISTS, and FVMD,
a perception--distortion trade-off in which distillation favors smoother,
more conservative outputs.
On Seoul, over 900-frame rollouts, OneFixer leads every metric except lane
jitter and Warp L1, and the two distilled baselines land close to
OmniDreams~@1, the diffusion-forcing fine-tune of the released DMD checkpoint.
Fig.~\ref{fig:computation_efficiency}(b) shows that OneFixer keeps the
lowest position-wise FVD throughout the 900-frame rollout rather than
gaining its aggregate advantage from a subset of the sequence, and
Fig.~\ref{fig:qualitative_comparison} shows representative cases of
improved local fidelity and preserved scene structure under 3DGS artifacts.

\paragraph{Training and inference efficiency.}
Fig.~\ref{fig:computation_efficiency}-(a) shows that OneFixer is
substantially faster than existing video-based fixers and approaches the
latency of image-based enhancement, since a single flow step per causal
block yields several RGB frames at once through the chunked causal decoder,
amortizing the backbone cost over the block.
Fig.~\ref{fig:computation_efficiency}-(c) compares training trajectories
against the same-backbone Self-Forcing and One-Forcing baselines, whose
three stages appear as regime changes in the curves.
OneFixer falls below both baselines' final FVD after roughly 345~GPU-hours,
before either baseline has completed teacher and causal adaptation, and
reaches the Self-Forcing FVMD plateau at the same point; it ends below both
baselines on both metrics.
The DMD stage delivers the expected one-step gain but saturates within its
first checkpoint, and a further 300--400~GPU-hours change either metric only
marginally; adding a One-Forcing critic~\citep{feng2026oneforcing} raises
the cost per iteration without closing the gap.
The plateau is therefore a ceiling of the distillation recipe rather than of
its budget, and since backbone, conditioning, and causal initialization are
held fixed, the difference is attributable to shared-rollout training.

\paragraph{Closed-loop replay fidelity.}
We drive a vision-language-action (VLA) policy in closed loop on 60
proprietary scenes with 5 seeds each, varying only the renderer
(Table~\ref{tab:closed-loop-eval}).
OneFixer yields the lowest collision rate, off-road rate, and deviation from
the logged route, reducing collisions by a third relative to the raw 3DGS
rendering (6.9\% vs.\ 10.6\%) and halving the off-road rate with
non-overlapping 95\% intervals.
The comparison is informative about what the policy needs: the image-based
DiFix improves the off-road rate but increases collisions relative to raw 3DGS, suggesting that improved frame-level appearance alone does not necessarily translate to policy-relevant consistency. The same-backbone OmniDreams and Self-Forcing stay
close to the unrefined baseline.
Because the scenes overlap with the policy's training data, these numbers
measure replay fidelity rather than policy generalization.
Fig.~\ref{fig:closed_loop_qualitative} shows a representative case: where the
3DGS rendering degrades, OneFixer's sharper, structurally faithful
observation is the only one under which the policy keeps following the road.

\subsection{Ablation and Analysis}
\label{sec:ablation}

Table~\ref{tab:onefixer-waymo-ablation} compares variants V1--V9, all
initialized from the same pre-DMD OmniDreams causal checkpoint and evaluated
at one step; V9 is the full model, SF denotes self-rollout flow matching,
and $k$ is the perceptual backpropagation horizon in latent positions.

\paragraph{Structured source.}
Replacing the opacity-mixed source with noise and channel-concatenated 3DGS
latents (V6) degrades every metric substantially, confirming that the
structured source is what makes one-step fine-tuning practical.

\paragraph{Shared rollout: objectives, coupling, and history.}
Self-rollout flow matching alone (V2) mainly improves temporal consistency
over V1 while leaving a perceptual gap; LPIPS alone (V3) improves fidelity
but is less consistent; combining both (V9) gives the best balance
(Fig.~\ref{fig:self_rollout_qualitative}).
How they are combined matters: sequential training under the same budget
(V7) matches V9 on frame-level fidelity but loses on FVD and FVMD, and
shortening the perceptual horizon to $k{=}1$ (V4) slightly improves LPIPS
while worsening both.
What the rollout contains matters as well: replacing the model's own
predictions with re-denoised ground truth as in
self-resampling~\citep{guo2026endend} (V8) degrades every metric, most in
LPIPS (0.158 vs.\ 0.117), since leaked clean content lets the model rely on
context it will not have at deployment.
Together these support optimizing both objectives jointly on the model's
own deployment-time rollout, so that preceding predictions are trained for
their role as future context.

\paragraph{Geometry conditioning.}
Removing lane and actor conditioning (V5) lowers Cuboid F1 and modestly
worsens perceptual and temporal metrics, yet V5 still beats every one-step
baseline in Table~\ref{tab:fixer-vs-baselines} on FVD, LPIPS, and Cuboid F1
with competitive FVMD, so the gains do not hinge on conditioning.
Robustness to lane and actor conditioning errors is evaluated in
Appendix~\ref{app:sdmap_robustness}.

\section{Conclusion and Limitations}

We introduced OneFixer, a one-step autoregressive video-diffusion fixer for
3DGS refinement in driving scenes.
Training the deployment-time generator on a single shared rollout, as both
history for flow matching and source for perceptual supervision, yields a
single-model, single-stage pipeline that delivers strong perceptual quality
and temporal consistency far beyond the training horizon at one-step latency.
Two limitations remain.
Fine structures such as text and road markings are not reliably recovered,
and strong perceptual scores do not guarantee their exactness.
Perceptual supervision through decoded predictions also makes peak training
memory grow with both the parallel temporal context and the perceptual
gradient window (Appendix~\ref{app:memory}); truncation keeps the default
configuration on a single device, and scaling to higher resolutions or more
cameras trades context for memory.
Cheaper feature-space supervision~\citep{wang2026seedvr2}, lightweight
decoders~\citep{zhuang2025flashvsr}, and reduced backbone
redundancy~\citep{guo2025oasis} are natural next steps, provided they
preserve causal refinement quality and the simplicity of shared-rollout
training.

\definecolor{bestblue}{RGB}{218,235,252}
\definecolor{secondgreen}{RGB}{222,242,224}
\definecolor{worstred}{RGB}{252,226,226}

\newcommand{\bestcell}[1]{\cellcolor{bestblue}#1}
\newcommand{\secondcell}[1]{\cellcolor{secondgreen}#1}
\newcommand{\worstcell}[1]{\cellcolor{worstred}#1}

\setlength{\dblfloatsep}{2pt}

\begin{table*}[t]
\centering
\caption{
Comparison on Waymo and proprietary Seoul dataset.
Denoising steps are denoted by @N.
Smaller raised values denote 95\% $t$-intervals over clips; FVD and FVMD are
single set-level Fr\'echet distances and therefore carry no per-clip interval.
For ArtiFixer, parentheses give the stage of its official pipeline at which
the model is evaluated (pre-DMD: after causal adaptation; DMD: after DMD with
Self-Forcing).
Harmonizer (temporal) conditions on its four previous outputs and is trained
with self-generated context (App.~\ref{app:baselines}).
OmniDreams, Self-Forcing, One-Forcing, and OneFixer share the OmniDreams
backbone and differ only in training recipe (Sec.~\ref{sec:exp_setup}).
Blue, green, and red cells indicate the best, second-best, and worst results
within each dataset, respectively.
}
\label{tab:fixer-vs-baselines}

\small
\setlength{\tabcolsep}{2.2pt}
\renewcommand{\arraystretch}{1.10}

\newcommand{\pmv}[2]{%
  #1{\raisebox{0.30ex}{\scalebox{0.85}{\tiny$\pm$#2}}}%
}

\begin{adjustbox}{width=\textwidth,center}
\begin{tabular}{lccccccccccc}
\toprule
&
\multicolumn{5}{c}{\textbf{Perception}}
&
\multicolumn{3}{c}{\textbf{Geometry}}
&
\multicolumn{3}{c}{\textbf{Temporal}}
\\
\cmidrule(lr){2-6}
\cmidrule(lr){7-9}
\cmidrule(lr){10-12}

Model
& FVD $\downarrow$
& LPIPS $\downarrow$
& DISTS $\downarrow$
& PSNR $\uparrow$
& SSIM $\uparrow$
& Cuboid F1 $\uparrow$
& Lane F1 $\uparrow$
& Lane $x$-err $\downarrow$
& Lane jit. $\downarrow$
& FVMD $\downarrow$
& Warp L1 $\downarrow$
\\
\midrule

\multicolumn{12}{c}{
\textbf{Waymo (80 scenes $\times$ 198 frames)}
} \\
\midrule

GSFix3D @8
& \worstcell{13.5}
& \pmv{0.182}{0.015}
& \pmv{0.185}{0.010}
& \pmv{22.9}{0.6}
& \worstcell{\pmv{0.781}{0.017}}
& \pmv{0.702}{0.058}
& \worstcell{\pmv{0.574}{0.074}}
& \pmv{0.178}{0.029}
& \worstcell{\pmv{0.167}{0.024}}
& \worstcell{285}
& \worstcell{\pmv{270}{23}}
\\

DiFix @1
& 13.4
& \pmv{0.142}{0.015}
& \pmv{0.126}{0.010}
& \pmv{23.6}{0.7}
& \pmv{0.810}{0.015}
& \pmv{0.776}{0.049}
& \pmv{0.708}{0.078}
& \pmv{0.136}{0.021}
& \pmv{0.094}{0.017}
& 195
& \pmv{123}{20}
\\

Harmonizer @1
& 10.5
& \pmv{0.136}{0.015}
& \pmv{0.124}{0.011}
& \secondcell{\pmv{24.5}{0.7}}
& \pmv{0.819}{0.015}
& \pmv{0.776}{0.050}
& \bestcell{\pmv{0.742}{0.076}}
& \pmv{0.127}{0.022}
& \pmv{0.093}{0.017}
& 179
& \pmv{115}{18}
\\

Harmonizer (temporal) @1
& 12.5
& \pmv{0.147}{0.015}
& \pmv{0.131}{0.010}
& \pmv{24.1}{0.7}
& \pmv{0.820}{0.014}
& \pmv{0.768}{0.050}
& \pmv{0.712}{0.077}
& \pmv{0.137}{0.025}
& \pmv{0.101}{0.019}
& 184
& \bestcell{\pmv{101}{17}}
\\

3DGS Enhancer @8
& 10.5
& \worstcell{\pmv{0.216}{0.019}}
& \worstcell{\pmv{0.211}{0.015}}
& \worstcell{\pmv{21.7}{0.6}}
& \pmv{0.784}{0.016}
& \worstcell{\pmv{0.609}{0.068}}
& \pmv{0.575}{0.075}
& \worstcell{\pmv{0.187}{0.027}}
& \pmv{0.099}{0.021}
& 259
& \pmv{112}{20}
\\

ArtiFixer (pre-DMD) @1
& 13.3
& \pmv{0.170}{0.019}
& \pmv{0.187}{0.012}
& \pmv{24.0}{0.7}
& \pmv{0.805}{0.016}
& \pmv{0.705}{0.056}
& \pmv{0.615}{0.081}
& \pmv{0.157}{0.028}
& \pmv{0.107}{0.024}
& 166
& \pmv{115}{21}
\\

ArtiFixer (pre-DMD) @4
& \secondcell{6.88}
& \pmv{0.147}{0.017}
& \pmv{0.145}{0.009}
& \pmv{23.4}{0.8}
& \pmv{0.797}{0.017}
& \pmv{0.728}{0.052}
& \pmv{0.636}{0.080}
& \pmv{0.155}{0.031}
& \pmv{0.108}{0.026}
& 165
& \pmv{116}{22}
\\

ArtiFixer (DMD) @1
& 8.46
& \pmv{0.146}{0.016}
& \pmv{0.158}{0.011}
& \pmv{24.1}{0.7}
& \pmv{0.811}{0.015}
& \pmv{0.744}{0.054}
& \pmv{0.675}{0.079}
& \pmv{0.142}{0.022}
& \pmv{0.104}{0.022}
& \bestcell{157}
& \pmv{119}{23}
\\

OmniDreams @1
& \worstcell{13.5}
& \pmv{0.184}{0.018}
& \pmv{0.177}{0.012}
& \pmv{23.7}{0.7}
& \pmv{0.829}{0.015}
& \pmv{0.758}{0.055}
& \pmv{0.645}{0.078}
& \pmv{0.134}{0.021}
& \pmv{0.094}{0.019}
& 216
& \pmv{125}{21}
\\

OmniDreams @4
& 11.9
& \secondcell{\pmv{0.132}{0.015}}
& \secondcell{\pmv{0.119}{0.008}}
& \pmv{23.6}{0.7}
& \pmv{0.829}{0.016}
& \secondcell{\pmv{0.791}{0.053}}
& \pmv{0.691}{0.078}
& \secondcell{\pmv{0.123}{0.020}}
& \pmv{0.093}{0.023}
& 172
& \pmv{119}{22}
\\

Self-Forcing @1
& 12.7
& \pmv{0.159}{0.017}
& \pmv{0.158}{0.011}
& \bestcell{\pmv{24.6}{0.7}}
& \secondcell{\pmv{0.838}{0.015}}
& \pmv{0.751}{0.058}
& \pmv{0.652}{0.078}
& \pmv{0.146}{0.034}
& \bestcell{\pmv{0.087}{0.015}}
& 175
& \secondcell{\pmv{102}{20}}
\\

One-Forcing @1
& 11.4
& \pmv{0.153}{0.017}
& \pmv{0.152}{0.011}
& \bestcell{\pmv{24.6}{0.7}}
& \bestcell{\pmv{0.839}{0.015}}
& \pmv{0.753}{0.058}
& \pmv{0.667}{0.078}
& \pmv{0.128}{0.027}
& \pmv{0.095}{0.017}
& 170
& \pmv{103}{20}
\\

\rowcolor{gray!12}
\textbf{OneFixer @1}
& \bestcell{5.83}
& \bestcell{\pmv{0.117}{0.013}}
& \bestcell{\pmv{0.101}{0.007}}
& \pmv{24.0}{0.7}
& \pmv{0.830}{0.015}
& \bestcell{\pmv{0.808}{0.052}}
& \secondcell{\pmv{0.726}{0.073}}
& \bestcell{\pmv{0.120}{0.019}}
& \secondcell{\pmv{0.088}{0.016}}
& \secondcell{158}
& \secondcell{\pmv{102}{21}}
\\
\midrule

\multicolumn{12}{c}{
\textbf{Proprietary Seoul (32 scenes $\times$ 900 frames)}
} \\
\midrule

GSFix3D @8
& 15.7
& \pmv{0.250}{0.011}
& \pmv{0.211}{0.011}
& \pmv{21.2}{0.6}
& \pmv{0.731}{0.024}
& \pmv{0.610}{0.072}
& \worstcell{\pmv{0.406}{0.047}}
& \worstcell{\pmv{0.368}{0.037}}
& \worstcell{\pmv{0.306}{0.028}}
& \worstcell{654}
& \worstcell{\pmv{301}{28}}
\\

DiFix @1
& \secondcell{7.78}
& \pmv{0.195}{0.011}
& \secondcell{\pmv{0.131}{0.004}}
& \worstcell{\pmv{19.3}{0.8}}
& \worstcell{\pmv{0.695}{0.024}}
& \secondcell{\pmv{0.742}{0.064}}
& \pmv{0.683}{0.029}
& \pmv{0.262}{0.024}
& \pmv{0.175}{0.011}
& 446
& \pmv{187}{26}
\\

Harmonizer @1
& \worstcell{30.7}
& \pmv{0.209}{0.014}
& \pmv{0.179}{0.009}
& \pmv{22.5}{0.6}
& \pmv{0.761}{0.023}
& \pmv{0.735}{0.063}
& \pmv{0.709}{0.034}
& \pmv{0.251}{0.024}
& \pmv{0.167}{0.011}
& 370
& \pmv{121}{18}
\\

Harmonizer (temporal) @1
& 28.0
& \pmv{0.244}{0.015}
& \pmv{0.211}{0.010}
& \pmv{21.4}{0.7}
& \pmv{0.756}{0.021}
& \pmv{0.720}{0.071}
& \pmv{0.652}{0.043}
& \pmv{0.281}{0.028}
& \pmv{0.153}{0.010}
& 423
& \bestcell{\pmv{99}{17}}
\\

3DGS Enhancer @8
& 18.9
& \worstcell{\pmv{0.282}{0.013}}
& \pmv{0.242}{0.009}
& \pmv{21.2}{0.6}
& \pmv{0.735}{0.024}
& \worstcell{\pmv{0.499}{0.071}}
& \pmv{0.537}{0.050}
& \pmv{0.337}{0.034}
& \bestcell{\pmv{0.137}{0.010}}
& 406
& \pmv{105}{17}
\\

ArtiFixer (DMD) @1
& 20.2
& \pmv{0.239}{0.014}
& \worstcell{\pmv{0.251}{0.011}}
& \pmv{20.6}{0.5}
& \pmv{0.724}{0.025}
& \pmv{0.671}{0.066}
& \pmv{0.614}{0.053}
& \pmv{0.366}{0.032}
& \pmv{0.145}{0.013}
& \secondcell{344}
& \pmv{111}{20}
\\

OmniDreams @1
& 27.4
& \pmv{0.251}{0.016}
& \pmv{0.241}{0.010}
& \pmv{22.1}{0.5}
& \secondcell{\pmv{0.770}{0.021}}
& \pmv{0.698}{0.067}
& \pmv{0.532}{0.071}
& \pmv{0.278}{0.032}
& \secondcell{\pmv{0.140}{0.015}}
& 432
& \pmv{134}{18}
\\

OmniDreams @4
& 13.1
& \secondcell{\pmv{0.193}{0.011}}
& \pmv{0.148}{0.005}
& \pmv{21.1}{0.6}
& \pmv{0.747}{0.022}
& \pmv{0.733}{0.067}
& \secondcell{\pmv{0.723}{0.030}}
& \secondcell{\pmv{0.242}{0.022}}
& \pmv{0.142}{0.012}
& 368
& \pmv{206}{29}
\\

Self-Forcing @1
& 27.2
& \pmv{0.238}{0.015}
& \pmv{0.239}{0.010}
& \secondcell{\pmv{22.7}{0.6}}
& \pmv{0.768}{0.022}
& \pmv{0.690}{0.065}
& \pmv{0.563}{0.058}
& \pmv{0.272}{0.032}
& \pmv{0.147}{0.017}
& 375
& \pmv{106}{19}
\\

One-Forcing @1
& 26.1
& \pmv{0.237}{0.015}
& \pmv{0.239}{0.010}
& \secondcell{\pmv{22.7}{0.6}}
& \pmv{0.769}{0.021}
& \pmv{0.690}{0.063}
& \pmv{0.564}{0.055}
& \pmv{0.272}{0.031}
& \pmv{0.149}{0.014}
& 369
& \pmv{106}{19}
\\

\rowcolor{gray!12}
\textbf{OneFixer @1}
& \bestcell{7.13}
& \bestcell{\pmv{0.148}{0.010}}
& \bestcell{\pmv{0.123}{0.006}}
& \bestcell{\pmv{22.9}{0.6}}
& \bestcell{\pmv{0.785}{0.021}}
& \bestcell{\pmv{0.766}{0.062}}
& \bestcell{\pmv{0.757}{0.029}}
& \bestcell{\pmv{0.220}{0.020}}
& \pmv{0.141}{0.012}
& \bestcell{312}
& \secondcell{\pmv{101}{18}}
\\
\bottomrule
\end{tabular}
\end{adjustbox}

\vspace{-7pt}
\end{table*}
\newcommand{\on}{\textcolor{green!55!black}{\ensuremath{\circ}}}
\newcommand{\off}{\textcolor{red!65!black}{\ensuremath{\times}}}
\newcommand{\onk}[1]{\textcolor{green!55!black}{\ensuremath{\circ_{#1}}}}
\newcommand{\tri}{\textcolor{green!55!black}{\ensuremath{\triangle}}}

\begin{table*}[t]
\centering

\newcommand{\pmv}[2]{%
  #1{\raisebox{0.30ex}{\scalebox{0.85}{\tiny$\pm$#2}}}%
}

\captionsetup[subtable]{font=footnotesize,skip=1pt}

\begin{subtable}[c]{0.58\textwidth}
\centering
\caption{Component ablation on Waymo.}
\label{tab:onefixer-waymo-ablation}

\scriptsize
\renewcommand{\arraystretch}{0.92}
\setlength{\tabcolsep}{0.8pt}


\begin{tabularx}{\linewidth}{
    @{}
    >{\centering\arraybackslash}p{0.055\linewidth}
    >{\centering\arraybackslash}p{0.050\linewidth}
    >{\centering\arraybackslash}p{0.026\linewidth}
    >{\centering\arraybackslash}p{0.070\linewidth}
    >{\centering\arraybackslash}p{0.055\linewidth}
    >{\centering\arraybackslash}p{0.055\linewidth}
    |
    >{\centering\arraybackslash}X
    >{\centering\arraybackslash}X
    >{\centering\arraybackslash}X
    >{\centering\arraybackslash}X
    @{}
}
\toprule
\textbf{ID} &
\multicolumn{5}{c|}{\textbf{Components}} &
\multicolumn{4}{c}{\textbf{Metrics}} \\
& SF & & LPIPS & Map & Mix &
\mbox{FVD $\downarrow$} &
LPIPS $\downarrow$ &
\mbox{Cuboid F1 $\uparrow$} &
FVMD $\downarrow$ \\
\midrule

V1 & \off & & \off & \on & \on
& 13.5 & \pmv{0.187}{0.018} & \pmv{0.758}{0.055} & 232 \\

V2 & \on & & \off & \on & \on
& 13.3 & \pmv{0.155}{0.017} & \pmv{0.759}{0.058} & 180 \\

V3 & \off & & \onk{4} & \on & \on
& \second{6.9} & \pmv{0.123}{0.014} & \pmv{0.795}{0.053} & 210 \\

V4 & \on & $+$ & \onk{1} & \on & \on
& 7.9 & \best{\pmv{0.114}{0.022}} & \pmv{0.793}{0.084} & 188 \\

V5 & \on & $+$ & \onk{4} & \off & \on
& 7.3 & \pmv{0.121}{0.013} & \pmv{0.787}{0.055} & \second{162} \\

V6 & \on & $+$ & \onk{4} & \on & \off
& 104.8 & \pmv{0.405}{0.028} & \pmv{0.715}{0.084} & 711 \\

V7 & \on & $\rightarrow$ & \onk{4} & \on & \on
& 7.7 & \pmv{0.119}{0.014} & \second{\pmv{0.803}{0.052}} & 193 \\

V8 & \tri & $+$ & \onk{4} & \on & \on
& 9.0 & \pmv{0.158}{0.018} & \pmv{0.785}{0.054} & 189 \\

\rowcolor{gray!12}
\textbf{V9} & \on & $+$ & \onk{4} & \on & \on
& \best{5.8} & \second{\pmv{0.117}{0.013}} & \best{\pmv{0.808}{0.053}} & \best{158} \\

\bottomrule
\end{tabularx}
\end{subtable}%
\hfill%
\begin{subtable}[c]{0.395\textwidth}
\centering
\caption{Closed-loop evaluation.}
\label{tab:closed-loop-eval}

\scriptsize
\renewcommand{\arraystretch}{0.95}
\setlength{\tabcolsep}{1.5pt}
\newcommand{\pms}[2]{#1{\raisebox{0.30ex}{\scalebox{0.85}{\tiny$\pm$#2}}}}
\begin{tabularx}{\linewidth}{
    @{}
    l
    >{\centering\arraybackslash}X
    >{\centering\arraybackslash}X
    >{\centering\arraybackslash}X
    @{}
}
\toprule
& \multicolumn{3}{c}{\textbf{Closed-loop Replay}} \\
Renderer &
Collision (\%) $\downarrow$ &
Route dist.\ (m) $\downarrow$ &
Off-road (\%) $\downarrow$ \\
\midrule

3DGS
& \pms{10.6}{2.2}
& \pms{4.18}{0.22}
& \pms{5.1}{1.3} \\

DiFix
& \pms{13.0}{2.4}
& \second{\pms{4.00}{0.19}}
& \second{\pms{3.3}{1.0}} \\

OmniDreams @4
& \second{\pms{10.2}{2.0}}
& \pms{4.31}{0.23}
& \pms{4.3}{1.2} \\

Self-Forcing @1
& \pms{10.7}{1.8}
& \pms{4.32}{0.22}
& \pms{3.6}{1.1} \\

\rowcolor{gray!12}
\textbf{OneFixer @1}
& \best{\pms{6.9}{1.7}}
& \best{\pms{3.96}{0.22}}
& \best{\pms{2.4}{0.9}} \\

\bottomrule
\end{tabularx}
\end{subtable}

\vspace{-3pt}

\caption{
\textbf{Ablation and closed-loop evaluation.}
(a) SF: causal history for flow matching --- \off{} ground truth, \tri{}
re-denoised ground truth (Sec.~\ref{sec:ablation}), \on{} own rollout.
\onk{k}: LPIPS with backpropagation horizon $k$ ($k{=}4$ default).
Map: lane and actor conditioning.
Mix: opacity-mixed source; \off{} uses noise with channel-concatenated 3DGS
latents.
$+$: joint; $\rightarrow$: 250 SF then 150 LPIPS-only steps.
(b) Closed-loop replay on 60 proprietary scenes $\times$ 5 policy seeds,
varying only the renderer; mean and 95\% interval over scene--seed runs.
Collision is the at-fault rate (front or lateral); route distance is the mean
deviation from the logged trajectory.
}
\label{tab:ablation-closedloop}

\vspace{-4pt}
\end{table*}

\section*{AI Use Statement}
Generative AI tools were used to assist with language editing, clarity,
and formatting of the manuscript. All technical content, experimental
results, and claims were reviewed and verified by the authors, who take
full responsibility for the final submission.

\section*{Reproducibility Statement}
Source code for training, inference, and the frozen evaluation protocol is
provided as supplementary material, including the exact Waymo recipe used
for Table~\ref{tab:fixer-vs-baselines} (\texttt{reproduce\_sv\_16gpu.sh}) and
the metric implementations described in App.~\ref{app:metrics}. The
pretrained checkpoint is the publicly released pre-DMD causal OmniDreams
model (App.~\ref{app:training}). Waymo data generation is described in
App.~\ref{app:data_waymo}; the Seoul dataset is proprietary and cannot be
released, but its preprocessing and training configuration are documented
in App.~\ref{app:data_internal} and \ref{app:training} and included in the
code. Baseline configurations follow App.~\ref{app:baselines}.

\bibliography{iclr2027_conference}

@article{rivkin2026chopgrad,
  title   = {ChopGrad: Pixel-Wise Losses for Latent Video Diffusion via Truncated Backpropagation},
  author  = {Rivkin, Dmitriy and Ewen, Parker and Gao, Lili and Ost, Julian and
             Walz, Stefanie and Kangutkar, Rasika and Bijelic, Mario and
             Heide, Felix},
  journal = {arXiv preprint arXiv:2603.17812},
  year    = {2026},
  doi     = {10.48550/arXiv.2603.17812}
}

@inproceedings{chen2024diffusionforcing,
  title        = {Diffusion Forcing: Next-token Prediction Meets Full-Sequence Diffusion},
  author       = {Chen, Boyuan and Mart\'i Mons\'o, Diego and Du, Yilun and Simchowitz, Max and Tedrake, Russ and Sitzmann, Vincent},
  booktitle    = {Advances in Neural Information Processing Systems},
  volume       = {37},
  year         = {2024}
}

@inproceedings{wu2025difix3d,
  title        = {{DIFIX3D+}: Improving {3D} Reconstructions with Single-Step Diffusion Models},
  author       = {Wu, Jay Zhangjie and Zhang, Yuxuan and Turki, Haithem and Ren, Xuanchi and Gao, Jun and Shou, Mike Zheng and Fidler, Sanja and Gojcic, Zan and Ling, Huan},
  booktitle    = {Proceedings of the IEEE/CVF Conference on Computer Vision and Pattern Recognition (CVPR)},
  pages        = {26024--26035},
  year         = {2025}
}

@inproceedings{fischerndartifixerenhancing,
  title        = {ArtiFixer: Enhancing and Extending {3D} Reconstruction with Auto-Regressive Diffusion Models},
  author       = {de Lutio, Riccardo and Fischer, Tobias and Chang, Yen-Yu and Zhang, Yuxuan and Wu, Jay Zhangjie and Ren, Xuanchi and Shen, Tianchang and Tothova, Katarina and Gojcic, Zan and Turki, Haithem},
  booktitle    = {ACM SIGGRAPH},
  year         = {2026}
}

@inproceedings{guo2026endend,
  title        = {End-to-End Training for Autoregressive Video Diffusion via Self-Resampling},
  author       = {Guo, Yuwei and Yang, Ceyuan and He, Hao and Zhao, Yang and Wei, Meng and Yang, Zhenheng and Huang, Weilin and Lin, Dahua},
  booktitle    = {European Conference on Computer Vision (ECCV)},
  year         = {2026}
}

@inproceedings{huangndselfforcing,
  title        = {Self Forcing: Bridging the Train-Test Gap in Autoregressive Video Diffusion},
  author       = {Huang, Xun and Li, Zhengqi and He, Guande and Zhou, Mingyuan and Shechtman, Eli},
  booktitle    = {Advances in Neural Information Processing Systems},
  volume       = {38},
  year         = {2025}
}

@article{li2026rollingsink,
  title        = {Rolling Sink: Bridging Limited-Horizon Training and Open-Ended Testing in Autoregressive Video Diffusion},
  author       = {Li, Haodong and Liu, Shaoteng and Lin, Zhe and Chandraker, Manmohan},
  journal      = {arXiv preprint arXiv:2602.07775},
  year         = {2026}
}

@article{nividiandcosmos3,
  title        = {Cosmos 3: Omnimodal World Models for Physical {AI}},
  author       = {{NVIDIA}},
  journal      = {arXiv preprint arXiv:2606.02800},
  year         = {2026}
}

@article{nvidia2026nvidiaomnidreams,
  title        = {{NVIDIA} OmniDreams: Real-Time Generative World Model for Closed-Loop Autonomous Vehicle Simulation},
  author       = {{NVIDIA} and Basant, Aarti and Kar, Amlan and Paschalidou, Despoina and Wei, Fangyin and Ferroni, Francesco and Cobo, Guillermo Garcia and Turki, Haithem and Ling, Huan and Seo, Jaewoo and Lucas, James and Wu, Jay Zhangjie and Wang, Jialiang and Lorraine, Jonathan and Gao, Jun and He, Kai and Tothova, Katarina and Xie, Kevin and Tyszkiewicz, Michał and Wu, Qi and de Lutio, Riccardo and Li, Ruilong and Fidler, Sanja and Kim, Seung Wook and Shen, Tianchang and Cao, Tianshi and Pfaff, Tobias and Lew, William and Wu, Xindi and Ren, Xuanchi and Lu, Yifan and Zhang, Yuxuan and Gojcic, Zan and Wang, Zian},
  journal      = {arXiv preprint arXiv:2606.03159},
  year         = {2026}
}

@article{wan2025wan,
  title        = {Wan: Open and Advanced Large-Scale Video Generative Models},
  author       = {Wan, Team and Wang, Ang and Ai, Baole and Wen, Bin and Mao, Chaojie and Xie, Chen-Wei and Chen, Di and Yu, Feiwu and Zhao, Haiming and Yang, Jianxiao and Zeng, Jianyuan and Wang, Jiayu and Zhang, Jingfeng and Zhou, Jingren and Wang, Jinkai and Chen, Jixuan and Zhu, Kai and Zhao, Kang and Yan, Keyu and Huang, Lianghua and Feng, Mengyang and Zhang, Ningyi and Li, Pandeng and Wu, Pingyu and Chu, Ruihang and Feng, Ruili and Zhang, Shiwei and Sun, Siyang and Fang, Tao and Wang, Tianxing and Gui, Tianyi and Weng, Tingyu and Shen, Tong and Lin, Wei and Wang, Wei and Wang, Wei and Zhou, Wenmeng and Wang, Wente and Shen, Wenting and Yu, Wenyuan and Shi, Xianzhong and Huang, Xiaoming and Xu, Xin and Kou, Yan and Lv, Yangyu and Li, Yifei and Liu, Yijing and Wang, Yiming and Zhang, Yingya and Huang, Yitong and Li, Yong and Wu, You and Liu, Yu and Pan, Yulin and Zheng, Yun and Hong, Yuntao and Shi, Yupeng and Feng, Yutong and Jiang, Zeyinzi and Han, Zhen and Wu, Zhi-Fan and Liu, Ziyu},
  journal      = {arXiv preprint arXiv:2503.20314},
  year         = {2025}
}

@article{wang2026diffusiondrf,
  title        = {Diffusion-{DRF}: Free, Rich, and Differentiable Reward for Video Diffusion Fine-Tuning},
  author       = {Wang, Yifan and Li, Yanyu and Qian, Gordon Guocheng and Tulyakov, Sergey and Fu, Yun and Kag, Anil},
  journal      = {arXiv preprint arXiv:2601.04153},
  year         = {2026}
}

@article{wu2026realitybridge,
  title        = {RealityBridge: Bridging Editable {3D} Gaussian Splatting Driving Simulations and Real-World Videos},
  author       = {Wu, Zhenhua and Pang, Yun and Chang, Mingkun and Ning, Yuwei and Wang, Liangzhi and Xiao, Yi and Li, Guanbin},
  journal      = {arXiv preprint arXiv:2606.16278},
  year         = {2026}
}

@inproceedings{liu20243dgs,
  title     = {3DGS-Enhancer: Enhancing Unbounded 3D Gaussian Splatting with View-Consistent 2D Diffusion Priors},
  author    = {Liu, Xi and Zhou, Chaoyi and Huang, Siyu},
  booktitle = {Advances in Neural Information Processing Systems},
  year      = {2024}
}

@inproceedings{zhou2024drivinggaussian,
  author    = {Zhou, Xiaoyu and Lin, Zhiwei and Shan, Xiaojun and
               Wang, Yongtao and Sun, Deqing and Yang, Ming-Hsuan},
  title     = {{DrivingGaussian}: Composite Gaussian Splatting for
               Surrounding Dynamic Autonomous Driving Scenes},
  booktitle = {Proceedings of the IEEE/CVF Conference on Computer Vision
               and Pattern Recognition (CVPR)},
  pages     = {21634--21643},
  year      = {2024}
}

@article{li2025mtgs,
  author        = {Li, Tianyu and Qiu, Yihang and Wu, Zhenhua and
                   Lindstr{\"o}m, Carl and Su, Peng and Nie{\ss}ner, Matthias
                   and Li, Hongyang},
  title         = {{MTGS}: Multi-Traversal Gaussian Splatting},
  journal       = {arXiv preprint arXiv:2503.12552},
  year          = {2025},
  eprint        = {2503.12552},
  archivePrefix = {arXiv}
}

@inproceedings{zhou2025flexdrive,
  author    = {Zhou, Jingqiu and Fan, Lue and Huang, Linjiang and
               Shi, Xiaoyu and Liu, Si and Zhang, Zhaoxiang and Li, Hongsheng},
  title     = {{FlexDrive}: Toward Trajectory Flexibility in Driving Scene
               Gaussian Splatting Reconstruction and Rendering},
  booktitle = {Proceedings of the IEEE/CVF Conference on Computer Vision
               and Pattern Recognition (CVPR)},
  pages     = {1549--1558},
  year      = {2025}
}

@inproceedings{sun2025splatflow,
  author    = {Sun, Su and Zhao, Cheng and Sun, Zhuoyang and
               Chen, Yingjie Victor and Chen, Mei},
  title     = {{SplatFlow}: Self-Supervised Dynamic Gaussian Splatting in
               Neural Motion Flow Field for Autonomous Driving},
  booktitle = {Proceedings of the IEEE/CVF Conference on Computer Vision
               and Pattern Recognition (CVPR)},
  pages     = {27487--27496},
  year      = {2025}
}

@article{zhou2026xiaomiauto,
  author        = {Zhou, Lijun and Luo, Hongcheng and Zhu, Zhenxin and
                   Chi, Cheng and Tu, Mingfei and Xiong, Kaixin and others},
  title         = {{Xiaomi EV World Model}: A Joint World Model Integrating
                   Reconstruction and Generation for Autonomous Driving},
  journal       = {arXiv preprint arXiv:2605.18137},
  year          = {2026},
  eprint        = {2605.18137},
  archivePrefix = {arXiv}
}

@inproceedings{yin2025slowbidirectional,
  title        = {From Slow Bidirectional to Fast Autoregressive Video Diffusion Models},
  author       = {Yin, Tianwei and Zhang, Qiang and Zhang, Richard and Freeman, William T. and Durand, Fredo and Shechtman, Eli and Huang, Xun},
  booktitle    = {Proceedings of the IEEE/CVF Conference on Computer Vision and Pattern Recognition (CVPR)},
  pages        = {22963--22974},
  year         = {2025}
}

@article{zhu2026causalforcing,
  title   = {Causal Forcing: Autoregressive Diffusion Distillation Done Right for High-Quality Real-Time Interactive Video Generation},
  author  = {Zhu, Hongzhou and Zhao, Min and He, Guande and Su, Hang and Li, Chongxuan and Zhu, Jun},
  journal = {arXiv preprint arXiv:2602.02214},
  year    = {2026},
  doi     = {10.48550/arXiv.2602.02214}
}

@article{zhang2026rewardforcing,
  title   = {Reward-Forcing: Autoregressive Video Generation with Reward Feedback},
  author  = {Zhang, Jingran and Li, Ning and Ban, Yuanhao and Bai, Andrew and Cui, Justin},
  journal = {arXiv preprint arXiv:2601.16933},
  year    = {2026},
  doi     = {10.48550/arXiv.2601.16933}
}

@article{zhang2026astrolabe,
  title   = {Astrolabe: Steering Forward-Process Reinforcement Learning for Distilled Autoregressive Video Models},
  author  = {Zhang, Songchun and Xue, Zeyue and Fu, Siming and Huang, Jie and Kong, Xianghao and Ma, Y and Huang, Haoyang and Duan, Nan and Rao, Anyi},
  journal = {arXiv preprint arXiv:2603.17051},
  year    = {2026},
  doi     = {10.48550/arXiv.2603.17051}
}

@inproceedings{yin2024dmd,
  title     = {One-step Diffusion with Distribution Matching Distillation},
  author    = {Yin, Tianwei and Gharbi, Micha{\"e}l and Zhang, Richard and
               Shechtman, Eli and Durand, Fr{\'e}do and Freeman, William T.
               and Park, Taesung},
  booktitle = {Proceedings of the IEEE/CVF Conference on Computer Vision
               and Pattern Recognition (CVPR)},
  pages     = {6613--6623},
  year      = {2024}
}

@article{you2026adaptivevideodistillation,
  author        = {You, Yuyang and Li, Yongzhi and Li, Jiahui and
                   Mu, Yadong and Chen, Quan and Jiang, Peng},
  title         = {Adaptive Video Distillation: Mitigating Oversaturation
                   and Temporal Collapse in Few-Step Generation},
  journal       = {arXiv preprint arXiv:2603.21864},
  year          = {2026},
  eprint        = {2603.21864},
  archivePrefix = {arXiv},
  primaryClass  = {cs.CV}
}

@article{feng2026oneforcing,
  author        = {Feng, Jiaqi and Cui, Justin and Ban, Yuanhao and
                   Hsieh, Cho-Jui},
  title         = {One-Forcing: Towards Stable One-Step Autoregressive
                   Video Generation},
  journal       = {arXiv preprint arXiv:2605.23458},
  year          = {2026},
  eprint        = {2605.23458},
  archivePrefix = {arXiv},
  primaryClass  = {cs.CV}
}

@article{kerbl2023gaussians,
  title   = {3D Gaussian Splatting for Real-Time Radiance Field Rendering},
  author  = {Kerbl, Bernhard and Kopanas, Georgios and Leimk{\"u}hler, Thomas and Drettakis, George},
  journal = {ACM Transactions on Graphics},
  volume  = {42},
  number  = {4},
  year    = {2023}
}

@inproceedings{yan2024street,
  title     = {Street Gaussians: Modeling Dynamic Urban Scenes with Gaussian Splatting},
  author    = {Yan, Yunzhi and Lin, Haotong and Zhou, Chenxu and Wang, Weijie
               and Sun, Haiyang and Zhan, Kun and Lang, Xianpeng
               and Zhou, Xiaowei and Peng, Sida},
  booktitle = {European Conference on Computer Vision (ECCV)},
  year      = {2024},
  doi       = {10.1007/978-3-031-73464-9_10}
}

@article{parmar2024onestep,
  title   = {One-Step Image Translation with Text-to-Image Models},
  author  = {Parmar, Gaurav and Park, Taesung and Narasimhan, Srinivasa and
             Zhu, Jun-Yan},
  journal = {arXiv preprint arXiv:2403.12036},
  year    = {2024}
}

@inproceedings{wu2024osediff,
  title     = {One-Step Effective Diffusion Network for Real-World Image
               Super-Resolution},
  author    = {Wu, Rongyuan and Sun, Lingchen and Ma, Zhiyuan and Zhang, Lei},
  booktitle = {Advances in Neural Information Processing Systems},
  volume    = {37},
  year      = {2024}
}

@article{zhang2024s3diff,
  title   = {Degradation-Guided One-Step Image Super-Resolution with
             Diffusion Priors},
  author  = {Zhang, Aiping and Yue, Zongsheng and Pei, Renjing and
             Ren, Wenqi and Cao, Xiaochun},
  journal = {arXiv preprint arXiv:2409.17058},
  year    = {2024}
}

@inproceedings{kim2026fidesr,
  title     = {{FiDeSR}: High-Fidelity and Detail-Preserving One-Step
               Diffusion Super-Resolution},
  author    = {Kim, Aro and Jang, Myeongjin and Moon, Chaewon and
               Shin, Youngjin and Jeong, Jinwoo and Park, Sang-hyo},
  booktitle = {Proceedings of the IEEE/CVF Conference on Computer Vision
               and Pattern Recognition (CVPR)},
  pages     = {38270--38280},
  year      = {2026}
}

@inproceedings{chen2025dove,
  title     = {{DOVE}: Efficient One-Step Diffusion Model for Real-World
               Video Super-Resolution},
  author    = {Chen, Zheng and Zou, Zichen and Zhang, Kewei and Su, Xiongfei
               and Yuan, Xin and Guo, Yong and Zhang, Yulun},
  booktitle = {Advances in Neural Information Processing Systems},
  volume    = {38},
  year      = {2025}
}

@inproceedings{wang2026seedvr2,
  title     = {{SeedVR2}: One-Step Video Restoration via Diffusion
               Adversarial Post-Training},
  author    = {Wang, Jianyi and Lin, Shanchuan and Lin, Zhijie and Ren, Yuxi
               and Wei, Meng and Yue, Zongsheng and Zhou, Shangchen and
               Chen, Hao and Zhao, Yang and Yang, Ceyuan and Xiao, Xuefeng
               and Loy, Chen Change and Jiang, Lu},
  booktitle = {International Conference on Learning Representations},
  year      = {2026}
}

@inproceedings{zhang2026diffusionharmonizer,
  title        = {DiffusionHarmonizer: Bridging Neural Reconstruction and Photorealistic Simulation with Online Diffusion Enhancer},
  author       = {Zhang, Yuxuan and T{\'o}thov{\'a}, Katar{\'\i}na and
                  Wang, Zian and Yin, Kangxue and Turki, Haithem and
                  de Lutio, Riccardo and Chang, Yen-Yu and Litany, Or and
                  Fidler, Sanja and Gojcic, Zan},
  booktitle    = {Proceedings of the IEEE/CVF Conference on Computer Vision and Pattern Recognition (CVPR)},
  year         = {2026}
}

@article{wei2025gsfix3d,
  title   = {{GSFix3D}: Diffusion-Guided Repair of Novel Views in {G}aussian Splatting},
  author  = {Wei, Jiaxin and Leutenegger, Stefan and Schaefer, Simon},
  journal = {arXiv preprint arXiv:2508.14717},
  year    = {2025}
}

@inproceedings{caesar2020nuscenes,
  title     = {nuScenes: A Multimodal Dataset for Autonomous Driving},
  author    = {Caesar, Holger and Bankiti, Varun and Lang, Alex H. and Vora, Sourabh and Liong, Venice Erin and Xu, Qiang and Krishnan, Anush and Pan, Yu and Baldan, Giancarlo and Beijbom, Oscar},
  booktitle = {Proceedings of the IEEE/CVF Conference on Computer Vision and Pattern Recognition},
  year      = {2020}
}

@inproceedings{sun2020scalability,
  title     = {Scalability in Perception for Autonomous Driving: Waymo Open Dataset},
  author    = {Sun, Pei and Kretzschmar, Henrik and Dotiwalla, Xerxes and Chouard, Aur{\'e}lien and Patnaik, Vijaysai and Tsui, Paul and Guo, James and Zhou, Yin and Chai, Yuning and Caine, Benjamin and others},
  booktitle = {Proceedings of the IEEE/CVF Conference on Computer Vision and Pattern Recognition},
  year      = {2020}
}

@article{hu2026autoawg,
  title   = {{AutoAWG}: Adverse Weather Generation with Adaptive Multi-Controls for Automotive Videos},
  author  = {Hu, Jiagao and Zhou, Daiguo and Fu, Danzhen and Li, Fuhao and
             Wang, Zepeng and Wang, Fei and Liao, Wenhua and Xie, Jiayi and
             Sun, Haiyang},
  journal = {arXiv preprint arXiv:2604.18993},
  year    = {2026}
}

@article{bai2026qwenvideoedit,
  title   = {Qwen-Video-Edit: Instruction-Based Video Editing by Repurposing an Image Editing Model},
  author  = {Bai, Yunpeng and Gandelsman, Yossi and Gharbi, Micha{\"e}l and Huang, Qixing},
  journal = {arXiv preprint arXiv:2608.14790},
  year    = {2026}
}

@inproceedings{wang2022pgd,
  author    = {Tai Wang and Xinge Zhu and Jiangmiao Pang and Dahua Lin},
  title     = {Probabilistic and Geometric Depth: Detecting Objects in Perspective},
  booktitle = {Proceedings of the 5th Conference on Robot Learning},
  series    = {Proceedings of Machine Learning Research},
  volume    = {164},
  pages     = {1475--1485},
  year      = {2022},
  publisher = {PMLR}
}

@inproceedings{luo2023latr,
  author    = {Yueru Luo and Chaoda Zheng and Xu Yan and Tang Kun and Chao Zheng
               and Shuguang Cui and Zhen Li},
  title     = {{LATR}: {3D} Lane Detection from Monocular Images with Transformer},
  booktitle = {Proceedings of the {IEEE/CVF} International Conference on Computer Vision ({ICCV})},
  pages     = {7941--7952},
  year      = {2023}
}

@inproceedings{chen2022persformer,
  author    = {Li Chen and Chonghao Sima and Yang Li and Zehan Zheng and Jiajie Xu
               and Xiangwei Geng and Hongyang Li and Conghui He and Jianping Shi
               and Yu Qiao and Junchi Yan},
  title     = {{PersFormer}: {3D} Lane Detection via Perspective Transformer and the {OpenLane} Benchmark},
  booktitle = {European Conference on Computer Vision ({ECCV})},
  pages     = {550--567},
  year      = {2022}
}

@article{liu2024fvmd,
  author  = {Jiahe Liu and Youran Qu and Qi Yan and Xiaohui Zeng and Lele Wang and Renjie Liao},
  title   = {Fr\'echet Video Motion Distance: A Metric for Evaluating Motion Consistency in Videos},
  journal = {arXiv preprint arXiv:2407.16124},
  year    = {2024}
}

@inproceedings{zheng2023pointodyssey,
  author    = {Yang Zheng and Adam W. Harley and Bokui Shen and Gordon Wetzstein and Leonidas J. Guibas},
  title     = {{PointOdyssey}: A Large-Scale Synthetic Dataset for Long-Term Point Tracking},
  booktitle = {Proceedings of the {IEEE/CVF} International Conference on Computer Vision ({ICCV})},
  pages     = {19855--19865},
  year      = {2023}
}

@inproceedings{farneback2003two,
  author    = {Gunnar Farneb{\"a}ck},
  title     = {Two-Frame Motion Estimation Based on Polynomial Expansion},
  booktitle = {Image Analysis: 13th Scandinavian Conference ({SCIA})},
  series    = {Lecture Notes in Computer Science},
  volume    = {2749},
  pages     = {363--370},
  year      = {2003},
  publisher = {Springer}
}

@misc{alpasim_2025,
  author = {{NVIDIA} and Yulong Cao and Riccardo de Lutio and Sanja Fidler
            and Guillermo Garcia Cobo and Zan Gojcic and Maximilian Igl
            and Boris Ivanovic and Peter Karkus and Janick Martinez Esturo
            and Marco Pavone and Aaron Smith and Ellie Tanimura
            and Michal Tyszkiewicz and Michael Watson and Qi Wu and Le Zhang},
  title  = {AlpaSim: A Modular, Lightweight, and Data-Driven Research
            Simulator for Autonomous Driving},
  year   = {2025},
  month  = {October},
  note   = {Software},
  url    = {https://github.com/NVlabs/alpasim}
}

@article{unterthiner2018towards,
  title={Towards Accurate Generative Models of Video: A New Metric \& Challenges},
  author={Unterthiner, Thomas and van Steenkiste, Sjoerd and Kurach, Karol and Marinier, Raphael and Michalski, Marcin and Gelly, Sylvain},
  journal={arXiv preprint arXiv:1812.01717},
  year={2018}
}

@InProceedings{Skorokhodov_2022_CVPR,
  author    = {Skorokhodov, Ivan and Tulyakov, Sergey and Elhoseiny, Mohamed},
  title     = {StyleGAN-V: A Continuous Video Generator With the Price, Image Quality and Perks of StyleGAN2},
  booktitle = {Proceedings of the IEEE/CVF Conference on Computer Vision and Pattern Recognition (CVPR)},
  month     = {June},
  year      = {2022},
  pages     = {3626-3636}
}

@article{luo2026policy,
  title={On-Policy Adversarial Flow Distillation for Autoregressive Video Generation},
  author={Luo, Yang and Qian, Shengju and Tang, Xiaohang and Zhu, Zirui and Liu, Yong and Wang, Xin and You, Yang},
  journal={arXiv preprint arXiv:2605.26105},
  year={2026}
}

@article{zheng2026causal,
  title={Causal-rCM: A Unified Teacher-Forcing and Self-Forcing Open Recipe for Autoregressive Diffusion Distillation in Streaming Video Generation and Interactive World Models},
  author={Zheng, Kaiwen and He, Guande and Zhao, Min and Zhang, Jintao and Chen, Huayu and Chen, Jianfei and Lin, Chen-Hsuan and Liu, Ming-Yu and Zhu, Jun and Ma, Qianli},
  journal={arXiv preprint arXiv:2606.25473},
  year={2026}
}

@article{zhao2026causal,
  title={Causal forcing++: Scalable few-step autoregressive diffusion distillation for real-time interactive video generation},
  author={Zhao, Min and Zhu, Hongzhou and Zheng, Kaiwen and Zhou, Zihan and Yan, Bokai and Li, Xinyuan and Yang, Xiao and Li, Chongxuan and Zhu, Jun},
  journal={arXiv preprint arXiv:2605.15141},
  year={2026}
}

@inproceedings{he2026ar,
  title={AR-CoPO: Align Autoregressive Video Generation with Contrastive Policy Optimization},
  author={He, Dailan and Feng, Guanlin and Ge, Xingtong and Zhang, Yi and Ma, Bingqi and Song, Guanglu and Liu, Yu and Li, Hongsheng},
  booktitle={European Conference on Computer Vision},
  pages={57--73},
  year={2026},
  organization={Springer}
}

@article{guo2025oasis,
  title   = {Towards Redundancy Reduction in Diffusion Models for Efficient Video Super-Resolution},
  author  = {Guo, Jinpei and Ji, Yifei and Chen, Zheng and Wang, Yufei and Ma, Sizhuo and Guo, Yong and Zhang, Yulun and Wang, Jian},
  journal = {arXiv preprint arXiv:2509.23980},
  year    = {2025}
}

@inproceedings{clark2024draft,
  title     = {Directly Fine-Tuning Diffusion Models on Differentiable Rewards},
  author    = {Clark, Kevin and Vicol, Paul and Swersky, Kevin and Fleet, David J.},
  booktitle = {International Conference on Learning Representations},
  year      = {2024}
}

@article{prabhudesai2023alignprop,
  title   = {Aligning Text-to-Image Diffusion Models with Reward Backpropagation},
  author  = {Prabhudesai, Mihir and Goyal, Anirudh and Pathak, Deepak and Fragkiadaki, Katerina},
  journal = {arXiv preprint arXiv:2310.03739},
  year    = {2023}
}

@article{zhuang2025flashvsr,
  title   = {{FlashVSR}: Towards Real-Time Diffusion-Based Streaming Video Super-Resolution},
  author  = {Zhuang, Junhao and Guo, Shi and Cai, Xin and Li, Xiaohui and Liu, Yihao and Yuan, Chun and Xue, Tianfan},
  journal = {arXiv preprint arXiv:2510.12747},
  year    = {2025}
}

@inproceedings{bai2022transfusion,
  title     = {{TransFusion}: Robust {LiDAR}-Camera Fusion for {3D} Object Detection with Transformers},
  author    = {Bai, Xuyang and Hu, Zeyu and Zhu, Xinge and Huang, Qingqiu and Chen, Yilun and Fu, Hongbo and Tai, Chiew-Lan},
  booktitle = {Proceedings of the IEEE/CVF Conference on Computer Vision and Pattern Recognition (CVPR)},
  pages     = {1090--1099},
  year      = {2022}
}
\bibliographystyle{iclr2027_conference}

\appendix

\section{Implementation Details}
\label{app:implementation}

\subsection{3DGS Data Generation - Waymo}
\label{app:data_waymo}
We construct a multi-modal driving corpus in which each sequence provides a
\emph{degraded} RGB render, the \emph{logged} front-camera RGB, an opacity map,
and a rasterized semantic map.
The objective is to train models that recover photometric fidelity and geometric
consistency when novel views are synthesized from an imperfect 3D scene
representation---a setting related to recent one-step diffusion refinements
of 3D reconstructions~\citep{wu2025difix3d}.
Our pairs are produced deterministically from explicit 3D Gaussian reconstructions
rather than from a diffusion model, yielding controllable and repeatable
degradation without generative stochasticity.

\paragraph{Reference reconstruction.}
Each sequence begins with a \emph{Street Gaussians}~\citep{yan2024street}
reconstruction of a dynamic urban clip: static background, sky, and moving actors
are modeled as separate Gaussian fields with time-varying actor poses, trained on
the logged camera trajectory with LiDAR depth and sky-mask supervision.
This reconstruction defines geometry, appearance, and actor motion for all
subsequent rendering.

\paragraph{Cycle-render degradation.}
To synthesize a structurally grounded but visually degraded input, we apply a
four-stage \emph{cycle render}:
\begin{enumerate}
  \item Render training views from the reference model with the camera shifted
        laterally by $1\,\mathrm{m}$ in the vehicle frame.
  \item Train a \emph{second} Street-Gaussian scene representation on these
        shifted views only.
        Actors are largely pruned during this stage because motion is inconsistent
        with static geometry in the shifted views.
  \item \emph{Splice} actor Gaussians and their pose tracks from the reference
        checkpoint into the cycle model so dynamic objects remain present at
        render time.
  \item Re-render at the \emph{logged} camera poses.
        Because the cycle model was trained $1\,\mathrm{m}$ to the left, this
        nominal-trajectory view corresponds to a $1\,\mathrm{m}$ right-shifted
        novel view relative to the cycle training cameras---introducing parallax,
        blur, and exposure mismatch while preserving coarse scene layout.
\end{enumerate}

The degraded RGB sequence forms \texttt{rgb\_input}; the original logged frames
form \texttt{rgb\_gt}.
Accumulated renderer alpha, with sky regions filled from segmentation masks,
forms \texttt{opacity}.
This lateral-extrapolation artifact is representative of novel-view errors from
feed-forward 3D representations and motivates learning-based restoration in the
spirit of reconstruction-oriented diffusion methods~\citep{wu2025difix3d}.

\paragraph{Semantic map (\texttt{sdmap}).}
Semantic-map videos are rasterized bird's-eye-view-style conditioning frames
($1920{\times}1280$ at 10\,fps) aligned to each sequence's front camera.
They follow the OmniDreams \emph{Ludus} HD-map renderer, which drives NVIDIA's
FlashDreams inference stack (\url{https://github.com/NVIDIA/flashdreams}) to
turn vector scene geometry into temporally consistent map video.
Concretely, per-frame \emph{clipGT} assets are built from public annotations:
OpenLane \texttt{lane3d} lane centerlines (white/yellow polylines and road
boundaries), together with 3D bounding boxes from the original dataset's native
object labels (laser-based detection boxes on dynamic actors and traffic signs
where available).
These vector pools and the logged camera trajectory are fed to
\texttt{LudusRenderer}, which produces stylized HD-map frames matched to each
timestamp.
In the cycle-render corpus, \texttt{sdmap} clips are taken from a precomputed
library generated with this pipeline rather than re-rendered at cycle time.
When Ludus is unavailable, an equivalent CPU rasterizer can project the same
OpenLane polylines and 3D boxes into the camera using standard calibration,
preserving semantic content at the cost of the learned Ludus appearance.
Map length may be shorter than the RGB stream on some sequences; in that case,
all modalities are truncated to the shortest available length before training.

\paragraph{Corpus statistics.}
Table~\ref{tab:cycle-render-stats} summarizes the constructed train/validation
splits.
Each scene contributes approximately 198 temporally aligned samples;
after truncation to the common length, the training split contains
\textbf{350} scenes and \textbf{69{,}238} frame-level pairs, and the validation
split contains \textbf{80} scenes and \textbf{15{,}812} pairs.
A smaller quality-filtered subset (verified actor splice and uniform frame counts)
contains 276 train and 59 validation scenes, of which 241 and 52 include
non-static traffic.
RGB and opacity are $1600{\times}1072$ at 24\,fps; semantic maps are
$1920{\times}1280$ at 10\,fps.

\begin{table}[t]
  \centering
  \caption{Cycle-render paired corpus statistics (front camera).}
  \label{tab:cycle-render-stats}
  \small
  \begin{tabular}{lrr}
    \toprule
    Quantity & Train & Validation \\
    \midrule
    Scenes (four modalities present) & 350 & 80 \\
    \quad with dynamic actors & 241 & 52 \\
    \midrule
    Median frames per scene & 198 & 198 \\
    Total aligned frame pairs & 69{,}238 & 15{,}812 \\
    \bottomrule
  \end{tabular}
\end{table}

\subsection{3DGS Data Generation - Internal}
\label{app:data_internal}
We use an internal multi-camera urban driving dataset collected from production
vehicles operating on public roads in South Korea, covering diverse urban and
suburban environments such as arterial roads, intersections, highways, and
dense city streets.
Each clip contains synchronized surround-view RGB streams with logged camera
calibration, ego-vehicle poses, and temporally aligned semantic-map metadata.
The original front-camera videos are recorded at $1920{\times}1080$ and are
processed at 10\,fps.
Training source clips use 30-second windows (300 frames), from which 93-frame
sequences are sampled; the 32 held-out evaluation scenes use merged 90-second
windows (900 frames) to test long-horizon rollouts.

We construct a paired driving-video corpus using multi-camera logged urban
sequences and 3D Gaussian scene reconstruction.
For each scene, we first build an MTGS scene representation following
\citet{li2025mtgs}, then render the trained representation to obtain a degraded
RGB sequence.
The rendered sequence is paired with the original logged RGB sequence at the
same camera trajectory, together with renderer opacity and a rasterized semantic
map.

\paragraph{Scene reconstruction.}
Each source clip is represented as a dynamic 3D Gaussian scene.
The reconstruction is trained from synchronized multi-camera video using the
logged camera calibration and trajectory.
We use the front, front-left, front-right, side-left, and side-right cameras
when available.
The training objective follows the MTGS/3DGS renderer stack: the Gaussian scene
is optimized to reproduce the logged views, while geometry-aware visibility is
captured through the rendered opacity channel.
Ego-vehicle regions are masked during preprocessing so that the reconstructed
scene models the external environment rather than the vehicle body.

\subsection{OneFixer Training Details}
\label{app:training}

\paragraph{Backbone and clip representation.}
We initialize OneFixer from the official OmniDreams causal checkpoint released
prior to DMD distillation.\footnote{
\url{https://huggingface.co/nvidia/omni-dreams-models/tree/main/single_view/student-init}}
This differs from the separately released DMD-distilled OmniDreams checkpoint
used for the OmniDreams baseline in Sec.~\ref{sec:results}.

Training uses $544\times960$ crops and 24 latent positions corresponding to
93 RGB frames.
Each causal block contains two latent positions, corresponding to 8 RGB frames
per generation.
The Waymo model uses a local-attention window of 12 latent positions; the
proprietary model uses a window of 6.
We do not use CFG at inference.

\paragraph{Self-rollout and truncated gradients.}
To supervise a self-generated rollout, we add a perceptual term on RGB
reconstructed from the model's own autoregressive history.
A target latent index is sampled uniformly from \(0\)--\(23\), covering the full
\(N{=}24\) clip.
The corresponding decoded frames (a few RGB frames per latent, following Wan's
temporal compression) are compared with identically decoded ground truth using
a frozen VGG LPIPS model, weighted approximately an order of magnitude above
the flow-matching term.

Backpropagation is truncated in two places.
On the autoregressive side, gradients are retained only through the last
\(4\) latent frames ending at the sampled index (two autoregressive blocks of
size \(2\)); earlier history is stop-gradient.
On the Wan decoder side, decoding is causal and cache-based, but the decoder
feature cache is detached at every latent frame.
Consequently, perceptual gradients do not propagate through decoder recurrence,
corresponding to a truncation distance of \(1\) following
\citet{rivkin2026chopgrad}.

\section{Further Analysis}
\label{app:analysis}

\subsection{Robustness to Geometry-Conditioning Errors}
\label{app:sdmap_robustness}

\paragraph{Evaluation protocol.}
To keep the computational cost tractable, we randomly sample
32 scenes from the 80-scene Waymo evaluation set and use
the same subset for all perturbation settings and reference
methods.
We perturb the lane and dynamic-agent geometry used to
construct the sdmap conditioning, while keeping the input
3DGS RGB renderings, opacity, and evaluation targets unchanged.
The perturbations are parameterized by a lane displacement
$\ell \in \{-0.50,-0.25,0,0.25,0.50\}$\,m and planar
object-box translations
$\delta_x,\delta_y \in \{-0.25,0,0.25\}$\,m.
The Cartesian sweep comprises 45 configurations.
For visualization, we group results by the absolute lane
displacement $|\ell|$ or object translation magnitude
$\sqrt{\delta_x^2+\delta_y^2}$, whose maximum is approximately
$0.354$\,m.
A zero value on one axis therefore does not necessarily
indicate fully unperturbed conditioning, since the other
parameters can remain nonzero.

\paragraph{Perturbation scale.}
The tested displacements are comparable in magnitude to
reported errors in driving-scene perception.
On OpenLane, LATR with a ResNet-50 backbone reports
near- and far-range lateral errors of approximately
$0.22$ and $0.26$\,m~\citep{luo2023latr};
our maximum lane displacement is approximately twice
these values.
For object geometry, TransFusion reports a mean
translation error of approximately $0.26$\,m on
nuScenes~\citep{bai2022transfusion}, comparable to the
$0.25$--$0.354$\,m nonzero translations tested here.
These benchmarks ground the sweep in representative
decimeter-scale errors.
The experiment is a controlled sensitivity test:
fixed geometric offsets do not reproduce the full
distribution of detection, localization, or map-alignment
errors encountered in deployment.

\begin{figure}[t]
    \centering
    \includegraphics[width=\linewidth]
    {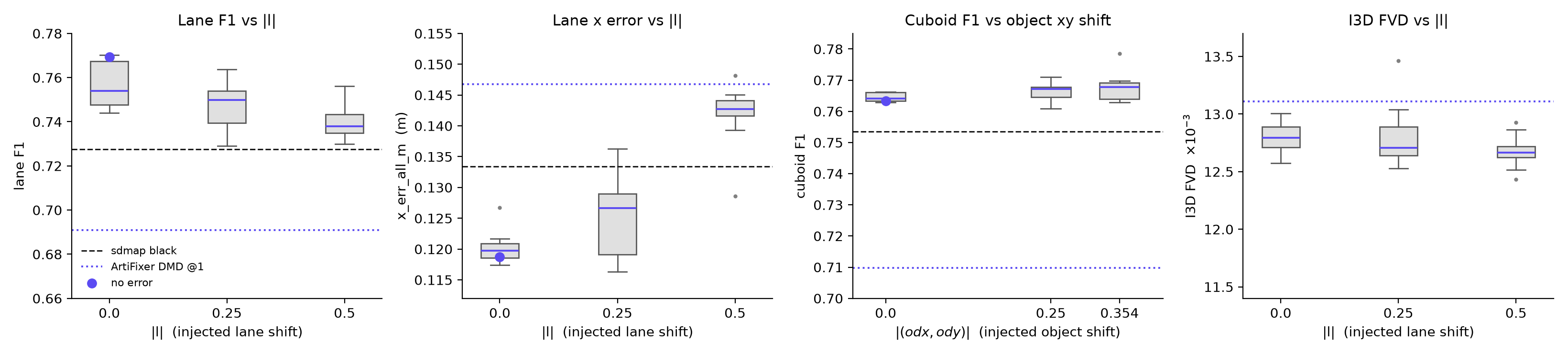}
    \caption{
    \textbf{Robustness to geometry-conditioning errors.}
    Evaluation uses a fixed random subset of 32 Waymo scenes.
    From left to right: Lane F1, lane lateral error,
    Cuboid F1, and I3D FVD, grouped by absolute lane
    displacement or planar object translation magnitude.
    Injected displacements are measured in meters.
    Boxplots summarize variation across perturbation
    configurations within each group.
    Black dashed lines indicate unconditioned OneFixer
    (black sdmap input), purple dotted lines indicate
    ArtiFixer DMD @1 on the same subset, and purple dots mark
    the fully unperturbed reference where shown.
    Lane F1 and Cuboid F1 remain above the unconditioned
    reference, although the largest lane displacement
    produces higher lateral error.
    }
    \label{fig:sdmap_robustness}
\end{figure}

\paragraph{Results.}
Fig.~\ref{fig:sdmap_robustness} shows that OneFixer
retains the benefits of geometry conditioning under
the tested perturbations.
Lane F1 remains above the unconditioned reference
throughout the sweep, while Cuboid F1 stays stable
across object translations and consistently exceeds
the unconditioned reference.
I3D FVD likewise shows no systematic degradation
with increasing lane displacement, suggesting that
overall video fidelity is robust to these errors.
Although lane lateral error exceeds the unconditioned
reference at the largest displacement, the retained
Lane F1 advantage indicates that conditioning remains
useful for recovering lane structure.
Overall, these results support the practical value
of imperfect geometric guidance, with a trade-off
in positional accuracy under larger lane misalignment.

\subsection{Training Memory}
\label{app:memory}

\begin{figure}[t]
    \centering
    \includegraphics[width=0.5\linewidth]{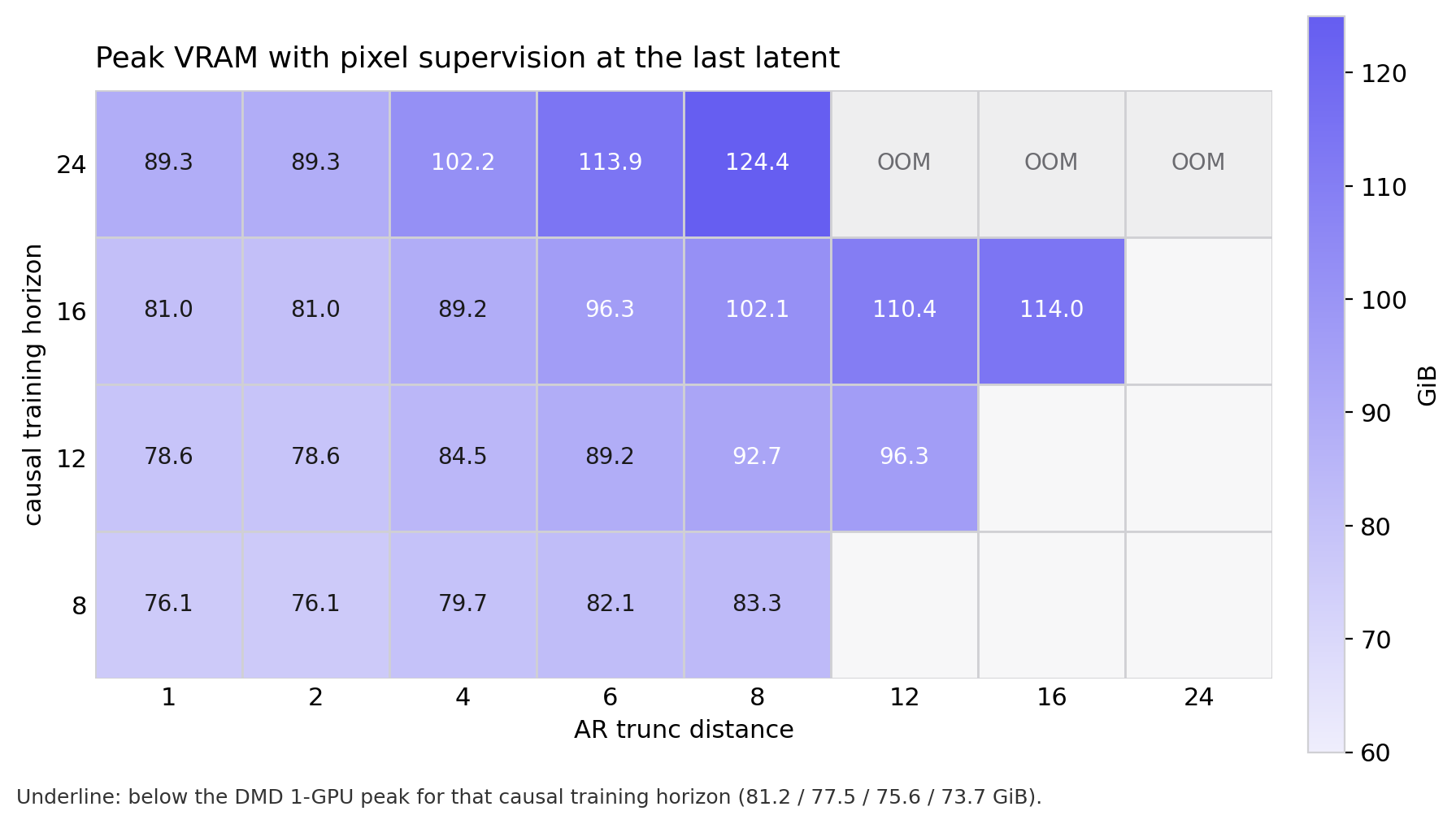}
    \caption{
    \textbf{Peak training memory versus parallel context and perceptual
    gradient window.}
    Single-GPU peak allocated memory (GiB) at $544\times960$ with one-step
    generation, a full rollout, and the perceptual loss on the final latent
    position, as a function of the number of latent positions processed in
    parallel (rows) and the perceptual backpropagation window $k$ in latent
    steps (columns).
    Grey cells exceed device capacity; blank cells have $k$ larger than the
    context.
    The default configuration (24 positions, $k=4$) is 102\,GiB.
    Multi-GPU training with sharded parameters shows comparable per-device
    peaks.
    }
    \label{fig:vram_heatmap}
\end{figure}

Direct perceptual supervision on the shared rollout retains gradients through
decoded predictions and the last $k$ latent positions of the autoregressive
history, so peak training memory depends on both the number of latent
positions processed in parallel and the perceptual gradient window.
Fig.~\ref{fig:vram_heatmap} measures single-GPU peak allocated memory at
$544\times960$ with the perceptual loss placed on the final latent, the
configuration that maximizes the retained history.
The default setting of 24 latent positions with $k=4$ requires about
102\,GiB, and memory increases monotonically along both axes: at 24 positions
it reaches 124\,GiB for $k=8$ and exceeds device capacity for $k\geq12$,
whereas shortening the parallel context to 16 positions accommodates $k$ up to
16.
Because each causal block spans two latent positions, $k=1$ and $k=2$ coincide.
Truncation therefore keeps the default configuration comfortably within a
single device, and the two knobs offer a direct trade-off: a longer perceptual
gradient window is available whenever a shorter parallel temporal context is
acceptable, which is the trade made for the multi-camera and high-resolution
settings in Appendix~\ref{sec:challenging_generation}.
This memory--horizon trade-off is not specific to OneFixer and also arises in
rollout-based autoregressive and distillation methods that optimize
self-generated trajectories.

\subsection{Generalization to Non-Driving Scenes.}
\begin{figure*}[t]
    \centering
    \includegraphics[width=\textwidth]{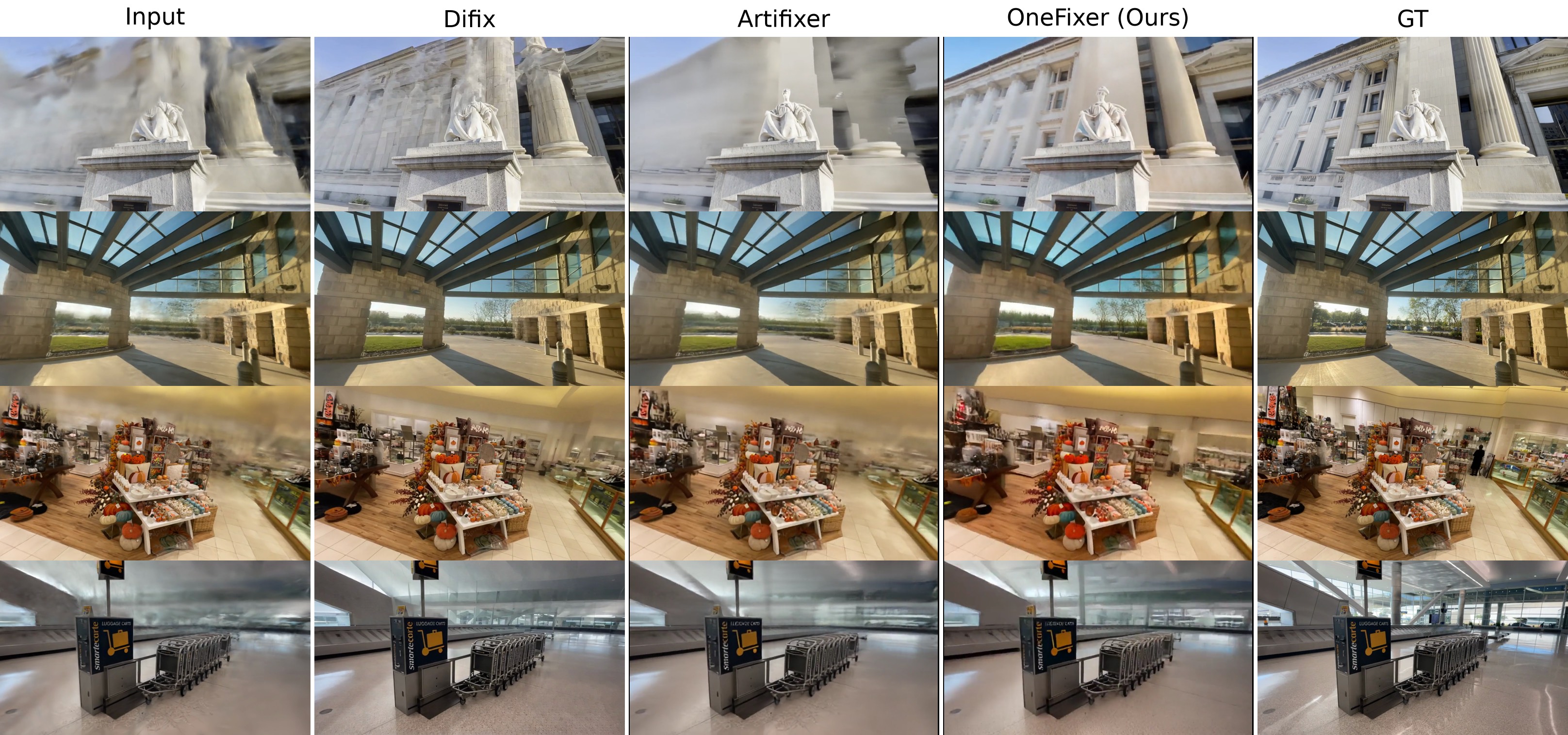}
    \caption{
    \textbf{Qualitative generalization on DL3DV.}
    OneFixer readily transfers beyond driving scenes after fine-tuning on DL3DV,
    while DiFix and ArtiFixer are evaluated using their official
    DL3DV-finetuned checkpoints.
    Under one-step inference, OneFixer matches or exceeds the frame-level
    visual quality of the image-based DiFix while providing stronger temporal
    consistency through video diffusion and produces higher-fidelity results
    than one-step ArtiFixer.
    }
    \label{fig:dl3dv}
\end{figure*}

To test transfer beyond driving, we fine-tune OneFixer on 32 DL3DV scenes
with degraded renders produced by under-trained 3dgs checkpoints, without lane or
actor conditioning (black sdmap input).
DiFix and ArtiFixer use their official DL3DV-finetuned checkpoints.
Table~\ref{tab:dl3dv-generalization} and Fig.~\ref{fig:dl3dv} report results
on the held-out scenes.

\section{Comparison}

\subsection{Evaluation Protocol and Metrics}
\label{app:metrics}

Unless otherwise specified, metrics are computed against the corresponding
clean reference sequence.
For metrics evaluated independently per clip, we report the mean together
with a 95\% $t$-interval across clips.
I3D FVD and FVMD are distributional metrics computed over an evaluation set
and therefore do not receive per-clip confidence intervals.
For compact presentation in Table~\ref{tab:fixer-vs-baselines}, I3D FVD,
FVMD, and Warp L1 are reported scaled by $10^{-3}$, $10^{-1}$, and $10^{4}$,
respectively.

\paragraph{FVD.}
We report I3D-based Fr\'echet Video Distance
(FVD)~\citep{unterthiner2018towards} using the full-clip evaluation protocol
adopted by StyleGAN-V~\citep{Skorokhodov_2022_CVPR}.
Unlike frame-wise perceptual metrics, FVD compares the distributions of
spatiotemporal video features extracted from generated and reference clips,
providing a complementary measure of overall video fidelity.
A single Fr\'echet distance is computed over the evaluation set.
Because the Fr\'echet estimate is biased upward for smaller sets and shorter
clips, FVD values on the 32-scene subsets used in
Fig.~\ref{fig:computation_efficiency}(c) (93-frame rollouts) and
Fig.~\ref{fig:sdmap_robustness} (198-frame rollouts) are higher than, and not
comparable to, the 80-scene, 198-frame values in
Table~\ref{tab:fixer-vs-baselines}, nor to each other; only within-figure
comparisons are meaningful there.

\paragraph{Perceptual metrics.}
We report LPIPS, PSNR, and SSIM for frame-level perceptual and reconstruction
quality.
Each metric is computed between generated and clean-reference frames and
aggregated over the evaluated sequence.
For evaluation, LPIPS uses the AlexNet backbone, following the common protocol
used in prior 3DGS enhancement benchmarks; this evaluation backbone is distinct
from the frozen VGG-based LPIPS used for training.
The reported value is the mean across clips, with the corresponding 95\%
$t$-interval.

\paragraph{DISTS.}
We additionally report DISTS, which compares deep-feature structure and
texture statistics and is less sensitive than PSNR to plausible high-frequency
detail that does not align pixel-wise with the reference.
We use the \texttt{pyiqa} implementation, compute the score per frame against
the clean reference, aggregate over the evaluated sequence, and report the mean
across clips with the corresponding 95\% $t$-interval.

\paragraph{Cuboid F1.}
We run a PGD monocular 3D detector~\citep{wang2022pgd} with a ResNet-101 FPN
and the nuScenes~\citep{caesar2020nuscenes} mono3D-finetuned checkpoint.
Images are rescaled so that the focal length matches 1266.42 pixels.
We retain nuScenes vehicle labels $\{0,1,2,3,4\}$ with score at least 0.10.
The same detector is run on generated and ground-truth videos; the
ground-truth reference is not derived from the map raster.

For each frame, a detected 7-DoF box is projected to its eight image-space
corners; boxes with any corner behind $z=0.5\,\mathrm{m}$ are removed.
We fill the projected convex hulls to obtain binary masks and aggregate
intersection, predicted area, and reference area over the clip.
With $P=I/A_{\mathrm{pred}}$ and $R=I/A_{\mathrm{ref}}$, we report
\begin{equation}
\mathrm{Cuboid\ F1}=\frac{2PR}{P+R}.
\end{equation}
Thus, Cuboid F1 measures projected-mask overlap rather than 3D-IoU box
matching.

\paragraph{Lane F1 and lateral error.}
We detect lanes with LATR~\citep{luo2023latr} using the
OpenLane~\citep{chen2022persformer} checkpoint, a class-score threshold of
0.50, at least three visible points, and at least $10\,\mathrm{m}$
longitudinal span.
Each lane is resampled at $y=3,5,\ldots,60\,\mathrm{m}$.
Lanes with less than $20\,\mathrm{m}$ support or
$|\operatorname{mean}x|>10\,\mathrm{m}$ are excluded.
Candidate pairs must overlap by at least $10\,\mathrm{m}$ and satisfy
$|\Delta x|\leq1.5\,\mathrm{m}$ for at least 75\% of their overlapping
samples.
We then perform Hungarian matching using mean absolute lateral error.
Lane F1 is computed from the resulting instance-level true positives,
false positives, and false negatives.
Lane lateral error is the mean $|\Delta x|$ over matched samples; the main
table reports the combined near-and-far value, with the boundary at
$30\,\mathrm{m}$.

\paragraph{Lane jitter.}
For a tracked lateral lane position $v_t$, we define the high-pass residual
\begin{equation}
h_t=v_t-\tfrac{1}{2}(v_{t-1}+v_{t+1}),
\end{equation}
and report the root mean square of valid residuals.
Lanes are associated across frames using a $1.5\,\mathrm{m}$ lateral
threshold and a minimum track length of five frames.
The reported lane jitter evaluates lateral position at
$y=40\,\mathrm{m}$; the evaluator additionally records a near-range value
at $y=20\,\mathrm{m}$.

\paragraph{FVMD.}
FVMD~\citep{liu2024fvmd} measures motion consistency using trajectory statistics
and complements appearance-based I3D FVD.
We compute FVMD on 16-frame windows sampled every 25 frames throughout each
evaluation sequence; thus, the reported score summarizes local motion
consistency across the full rollout rather than direct long-range consistency
between distant frames.
Each window is JPEG-encoded at quality 95, resized to $256\times256$, and
processed by PIPs++~\citep{zheng2023pointodyssey} to track 400 points with
stride 8.
The resulting trajectories are represented by velocity and acceleration
histograms using a $5\times5$ spatial grid, four-frame temporal subcubes, and
eight angle bins.
FVMD is the Fr\'echet distance between the generated and reference trajectory
feature distributions.

\paragraph{Flow-warped L1.}
Warp L1 measures short-term temporal consistency over the late evaluation
window.
We estimate dense optical flow from the current grayscale frame to the
previous frame using Farneb\"ack flow~\citep{farneback2003two}, with pyramid
scale 0.5, three levels, window size 21, three iterations, polynomial
neighborhood 5, and polynomial sigma 1.2.
The previous RGB frame is backward-warped with bilinear interpolation, and
samples whose warped coordinates fall outside the image are excluded.
When a valid road-region mask is available, the error is restricted to that
region; otherwise, all valid in-bounds pixels are used.
For valid pixels $\Omega$, we compute
\begin{equation}
\frac{1}{|\Omega|}\sum_{p\in\Omega}
\frac{
\left\lVert
I_t(p)-\mathcal{W}(I_{t-1})(p)
\right\rVert_1
}{3\cdot255}.
\end{equation}
The metric is averaged over the designated late temporal window for each
clip, and Table~\ref{tab:fixer-vs-baselines} reports the mean and 95\%
$t$-interval across clips.

\paragraph{Closed-loop replay evaluation.}
We evaluate closed-loop replay fidelity on 60 proprietary scenes, each run
with 5 policy seeds, using the AlpaSim evaluators~\citep{alpasim_2025}.
Each scene--seed run is repeated several times and averaged; a small number
of runs lost to system failures are excluded.
We use the same per-timestep scorers and aggregation as the public AlpaSim
evaluation: warm-up frames are discarded until the evaluator marks them as
relevant (\texttt{eval\_relevant}), and frames after the first off-road or
collision event are excluded.
Binary metrics are aggregated by taking the per-rollout $\mathrm{MAX}$,
averaging within each scene--seed run, and then averaging across runs;
results are reported as percentages with a 95\% interval
($1.96\,\sigma/\sqrt{N}$) over the $N$ runs.
The ego footprint is represented by a centered axis-aligned bounding box of
size $4.855\times1.88\,\mathrm{m}$, without shrinkage or corner rounding.

\paragraph{Collision.}
At each step, we construct bird's-eye-view polygons for the ego vehicle and
all other actors and query intersections with an STRtree.
A colliding actor is classified as \emph{front} if it intersects the ego
front-bumper segment, \emph{rear} if it intersects the rear-bumper segment,
and \emph{lateral} otherwise.
We report AlpaSim's at-fault collision rate, the union of front and lateral
collisions; rear impacts are scored but not reported.

\paragraph{Route distance.}
At each relevant step we measure the distance (m) from the ego position to
the logged trajectory and average over the rollout; the table reports the
mean over runs.

\paragraph{Off-road.}
Off-road evaluation uses the vector lane map.
Nearby lanes are queried within $6\,\mathrm{m}$ of the ego vehicle without
a heading constraint.
A step is considered on-road if any individual lane polygon fully contains
the ego footprint or, when the vehicle straddles adjacent lanes, if the union
of nearby lane polygons contains it.
Lane polygons are constructed from left and right lane edges when both are
available; otherwise, a $1.85\,\mathrm{m}$ half-width buffer around the
centerline is used.
If neither containment test passes, the step is classified according to its
distance to the nearest road-edge polyline using the evaluator's
$0.001\,\mathrm{m}$ threshold.
The reported value is the mean over runs of the per-rollout $\mathrm{MAX}$
of this binary indicator.

\subsection{Baseline Configuration}
\label{app:baselines}

All baseline videos are evaluated using the same frozen protocol described in
App.~\ref{app:metrics}.
We report each method at the operating point used in the main comparison rather
than forcing all methods to use the same number of denoising steps.

\paragraph{GSFix3D.}
We initialize from the official GSFix3D checkpoint~\citep{wei2025gsfix3d}
(\texttt{gsfixer-base}) and fine-tune the UNet on paired degraded/clean videos
from the internal dataset and Waymo at $544\times960$.
Training uses 8 GPUs, per-GPU batch 2, and gradient accumulation 4
(effective batch 64), Adam at $3\times10^{-5}$, bfloat16, and 4 denoising
steps.
The reported checkpoint is iteration 10{,}000.
The table uses 8-step inference.

\paragraph{DiFix.}
We initialize from the official NVIDIA DiFix checkpoint~\citep{wu2025difix3d}
(\texttt{nvidia/difix}) and fine-tune on the same internal dataset and Waymo.
Training uses learning rate $2\times10^{-5}$, per-GPU batch 4, accumulation 2,
bfloat16, and L2 + LPIPS + Gram losses (Gram warmup 2{,}000 steps).
The reported run is 800 optimization steps.
Inference is independent per frame at diffusion timestep 199 with prompt
\texttt{remove degradation}: generation is performed at $576\times1024$ and
then resized to $544\times960$.
Evaluation uses one-step inference.

\paragraph{Harmonizer.}
We initialize from the official DiffusionHarmonizer
checkpoint~\citep{zhang2026diffusionharmonizer} (\texttt{nvidia/Harmonizer})
and fine-tune the full network at $544\times960$ on the internal dataset and
Waymo for 4k steps each, following the released training command: 8 GPUs,
per-GPU batch 4, accumulation 2 (effective batch 64), learning rate
$2\times10^{-5}$, bfloat16, timestep 250, $\lambda_{\mathrm{L2}}=1$,
$\lambda_{\mathrm{LPIPS}}=0.3$.
\emph{Harmonizer} (frame-wise) trains and infers on single frames.
\emph{Harmonizer (temporal)} continues from the frame-wise checkpoint for
1{,}500 further steps and conditions on the four previously enhanced frames
at inference ($K{=}4$, offsets $-1$ to $-4$), as in the official inference
script.
The released temporal training path conditions on ground-truth prior frames
(with 50\% context dropout) and applies a RAFT warping loss; on our
sequences this teacher-forced recipe drifted within a few dozen frames when
rolled out on its own outputs.
We therefore trained the temporal variant with self-generated context: the
four references are the model's own detached, re-encoded outputs from a
short autoregressive window, with the warping loss retained.
This is the strongest temporal Harmonizer we could obtain on our data; it is
a modification of the official recipe, not a reproduction of the paper's
checkpoint.
Both variants are evaluated at one step.

\begin{figure}[t]
    \centering
    \includegraphics[width=\linewidth]{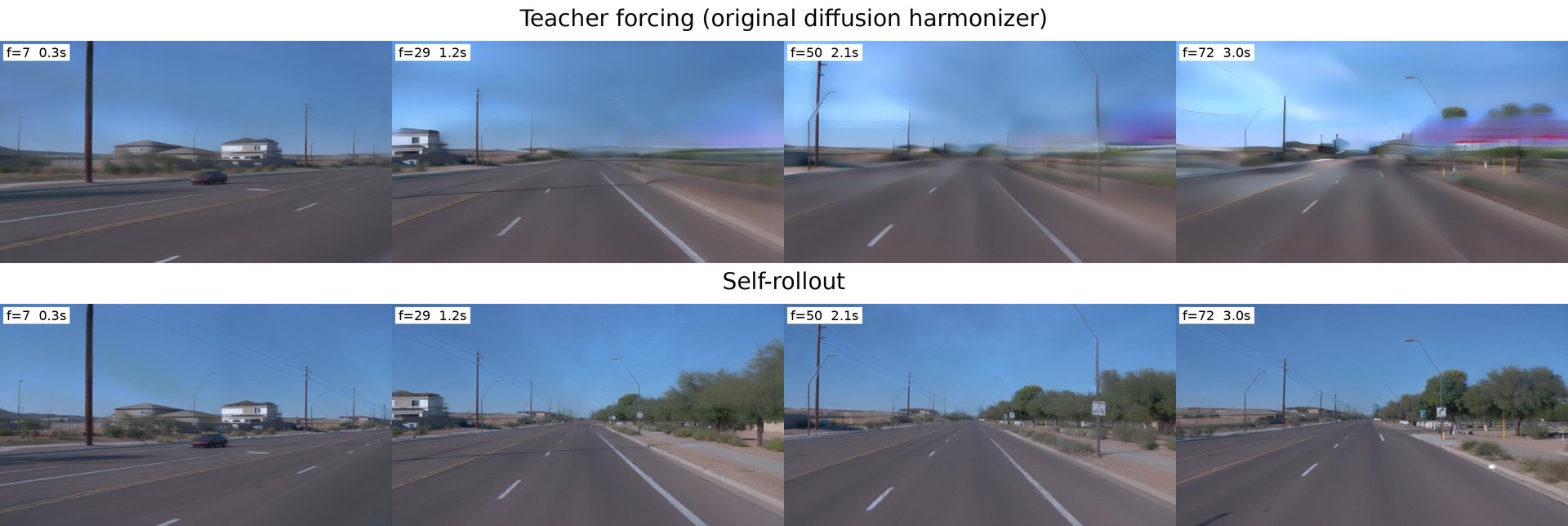}
    \caption{
    \textbf{Exposure bias in a frame-recurrent image fixer.}
    Harmonizer in temporal mode on a Waymo sequence, conditioning on its own
    four previous outputs at inference.
    Top: trained with the official teacher-forced recipe (ground-truth prior
    frames as context, 50\% context dropout); appearance drifts within a few
    dozen frames and the model begins copying its own artifacts.
    Bottom: the same model trained with self-generated context; the rollout
    remains stable.
    This is the variant reported as Harmonizer (temporal) in
    Table~\ref{tab:fixer-vs-baselines}.
    Frame index and elapsed time are shown in each panel.
    }
    \label{fig:harmonizer_drift}
\end{figure}

\paragraph{3DGS Enhancer.}
We initialize from Stable Video Diffusion (\texttt{img2vid-xt}) and fine-tune
3DGS-Enhancer~\citep{liu20243dgs}, a \emph{bidirectional} (non-causal) video
diffusion UNet.
Fine-tuning uses 25-frame clips at $640\times960$ on the internal dataset and
Waymo, with Adam at $1\times10^{-5}$, weight decay $0.01$, fp16, batch 1 per
GPU, and 500 steps.
Outputs are resized to $544\times960$ for evaluation.
Because the backbone is bidirectional and clip-length limited, full evaluation
sequences are produced using overlapping 25-frame windows (overlap 8) and
stitching.
Guidance is 1.0.
The main comparison uses 8-step inference.

\paragraph{ArtiFixer.}
We initialize from the official ArtiFixer 1.3B checkpoint
(\texttt{artifixer-1.3b})~\citep{fischerndartifixerenhancing}, built on
Wan2.1-T2V-1.3B~\citep{wan2025wan}, and follow the three-stage pipeline on the
internal dataset and Waymo at $544\times960$ using 93-frame clips.
Training uses 16 GPUs with accumulation 4 (effective batch 64):
400 steps of bidirectional teacher training, 600 steps of causal adaptation
($3\times10^{-5}$), and 200 steps of DMD/Self-Forcing
(generator $2\times10^{-6}$, critic $4\times10^{-7}$, EMA 0.99).
The reported result is the one-step DMD model with text guidance 1.0.

\paragraph{OmniDreams.}
We initialize from the official DMD-distilled causal OmniDreams
checkpoint~\citep{nvidia2026nvidiaomnidreams}
(\url{https://huggingface.co/nvidia/omni-dreams-models/tree/main/single_view/distilled})
and fine-tune it with diffusion forcing and the opacity-mixing rule on the
internal dataset and Waymo.
In contrast, OneFixer starts from the separately released causal checkpoint
prior to DMD distillation.
Training uses 8 GPUs, per-GPU batch 1, and accumulation 8
(effective batch 64) for 400 iterations at $544\times960$ using 93-frame
windows.
The main comparison uses the 4-step operating point with guidance 1.0.

\paragraph{Self-Forcing and One-Forcing.}
Following the staged recipe in Sec.~\ref{sec:exp_setup}, the bidirectional
teacher is fine-tuned from the official pre-DMD bidirectional OmniDreams
checkpoint~\citep{nvidia2026nvidiaomnidreams} to 3DGS refinement
(Stage~1); the causal student is fine-tuned from the official pre-DMD
causal checkpoint, OneFixer's initialization, to the fixer task (Stage~2);
the Stage-2 student is distilled with DMD and
Self-Forcing~\citep{yin2024dmd,huangndselfforcing} against the Stage-1
teacher (Stage~3).
Stages use 400/400/400 steps on Waymo and 500/500/400 on Seoul at
$544\times960$, effective batch 64, with the same lane and dynamic-agent
conditioning as OneFixer.
One-Forcing adds the critic module and loss of \citet{feng2026oneforcing} to
the third stage, which raises the per-iteration cost and accounts for its
higher cumulative GPU-hours.
Both are evaluated at one denoising step.

\begin{table}[t]
\centering
\caption{
Generalization to non-driving scenes on DL3DV.
Denoising steps are denoted by @N.
Best and second-best results are shown in
\textbf{bold} and \underline{underlined}, respectively.
FVMD is reported unscaled here.
}
\label{tab:dl3dv-generalization}

\small
\setlength{\tabcolsep}{4.0pt}
\renewcommand{\arraystretch}{1.10}

\begin{tabular}{lccccc}
\toprule
&
\multicolumn{2}{c}{\textbf{Temporal}}
&
\multicolumn{3}{c}{\textbf{Perception}}
\\

\cmidrule(lr){2-3}
\cmidrule(lr){4-6}

Model
&
Warp L1 $\downarrow$
&
FVMD $\downarrow$
&
LPIPS-Mid $\downarrow$
&
LPIPS-Last $\downarrow$
&
PSNR $\uparrow$
\\
\midrule

DiFix @1
&
0.0306
&
6744
&
\second{0.140}
&
\second{0.164}
&
21.5
\\

ArtiFixer DMD @4
&
\best{0.0179}
&
\second{3010}
&
0.148
&
\second{0.164}
&
22.2
\\

ArtiFixer DMD @1
&
\second{0.0181}
&
3411
&
0.176
&
0.193
&
\second{22.3}
\\

\rowcolor{gray!12}
\textbf{OneFixer (Ours) @1}
&
0.0201
&
\best{2326}
&
\best{0.126}
&
\best{0.136}
&
\best{22.6}
\\

\bottomrule
\end{tabular}
\end{table}

\section{Appearance-Conditioned Refinement}
\label{app:stylization}

\begin{figure*}[t]
    \centering
    \includegraphics[width=\textwidth]{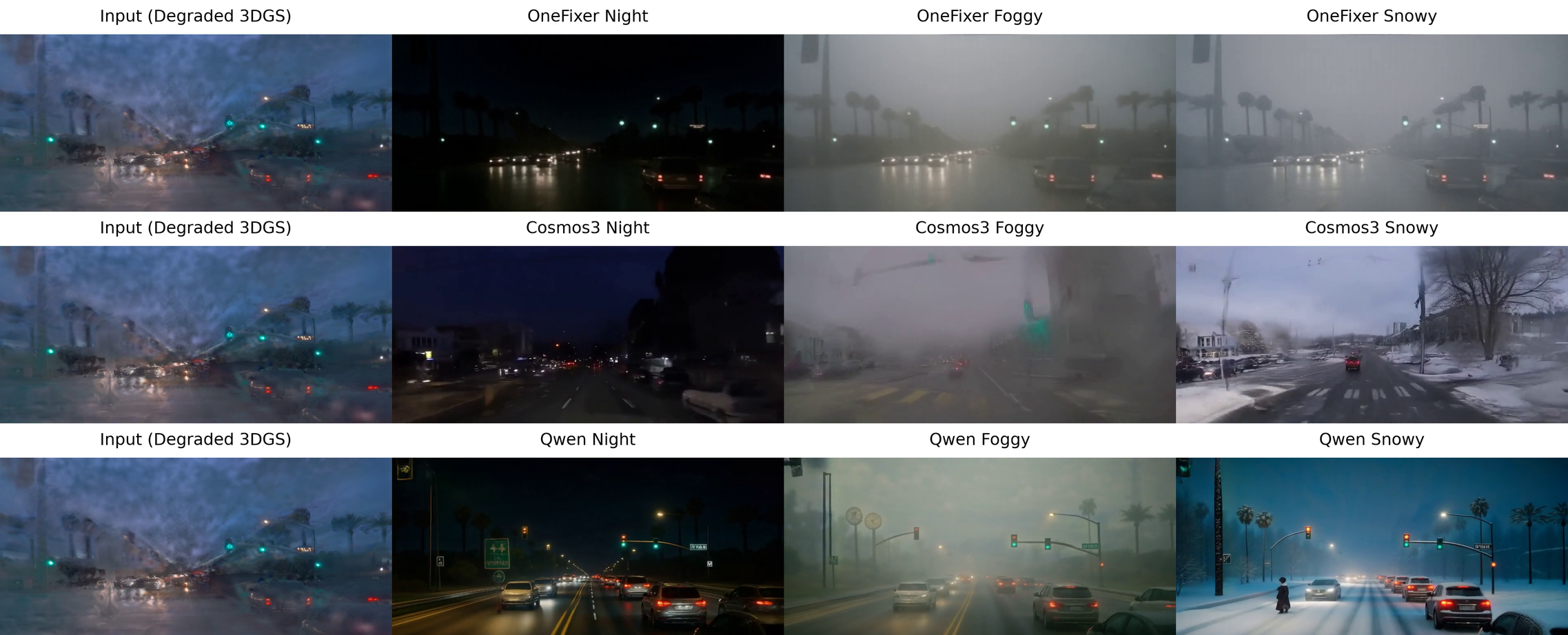}
    \caption{
    \textbf{Appearance-conditioned 3DGS refinement.}
    Given the same degraded 3DGS rendering, we compare prompt-conditioned
    generation for night, fog, and snow.
    General-purpose video editing models such as Cosmos3 and Qwen-Video-Edit
    can produce visually convincing target styles, but often substantially
    alter the underlying scene content and geometry inherited from the 3DGS
    input.
    In contrast, after paired prompt-conditioned training, OneFixer jointly
    performs 3DGS refinement and appearance transformation in a single pass,
    removing reconstruction artifacts while better preserving the underlying
    scene structure.
    }
    \label{fig:stylization}
\end{figure*}

We further investigate whether OneFixer can extend beyond clear-weather
restoration to \emph{appearance-conditioned 3DGS refinement}.
To construct paired supervision without requiring matched captures under rare
conditions, we use AutoAWG~\citep{hu2026autoawg} to transform the clean
reference videos from our Waymo cycle-render training set into three target
conditions: \emph{night}, \emph{fog}, and \emph{snow}.
The generated targets preserve the original camera trajectory and scene
content, and outputs exhibiting noticeable structural drift are filtered using
alignment with the corresponding clean sequence.
The original clear-weather sequence is retained as an identity condition.
Importantly, the input to OneFixer is unchanged across styles: the model always
receives the degraded, viewpoint-shifted 3DGS rendering together with its
opacity and HD-map conditioning.

We initialize from the same official OmniDreams causal checkpoint released
prior to DMD distillation~\citep{nvidia2026nvidiaomnidreams} and post-train
OneFixer on an equal mixture of identity, night, fog, and snow targets.
The requested appearance is provided through text conditioning, while training
otherwise follows the same shared-rollout objective used by the clear-weather
model, including self-rollout flow matching and direct perceptual supervision.
Consequently, the model is not trained as a conventional video stylizer that
operates on an already clean input.
Instead, it must jointly interpret the structure implied by an imperfect 3DGS
rendering, correct reconstruction artifacts, and synthesize the requested
appearance in a single end-to-end refinement process.

We compare this capability with general-purpose video editing models,
including Cosmos~3~\citep{nividiandcosmos3} and
Qwen-Video-Edit~\citep{bai2026qwenvideoedit}.
While these models can produce strong appearance changes, they do not
consistently resolve reconstruction errors inherited from the underlying
3DGS rendering.
OneFixer instead combines structural restoration and appearance conditioning
within the same generative process, enabling the refined output to remain
consistent with the inferred scene structure while adapting its illumination
and weather appearance.
We evaluate target-style alignment and geometric fidelity to the corresponding
clean condition using complementary vision-language and recognition-based
metrics, aggregated across frames and scenes (Table~\ref{tab:stylization}).
These results suggest that a 3DGS fixer can serve not only as a restoration
module but also as an end-to-end interface for controllable rendering
enhancement.
Further improving fine-grained appearance control through broader and more
diverse training data remains an interesting direction for future work.

\begin{table}[t]
\centering
\caption{Style conditioning.}
\label{tab:stylization}
\begin{tabular}{lcccc}
\toprule
Model
& \multicolumn{2}{c}{Style}
& \multicolumn{2}{c}{Geometry} \\
\cmidrule(lr){2-3}
\cmidrule(lr){4-5}
& \shortstack{CLIP\\[-2pt]acc $\uparrow$}
& \shortstack{VLM\\[-2pt]acc $\uparrow$}
& \shortstack{Lane\\[-2pt]F1 $\uparrow$}
& \shortstack{Lane\\[-2pt]$x$-err $\downarrow$} \\
\midrule

Cosmos3
& \underline{0.962}
& \underline{0.992}
& 0.098
& 0.632 \\

Qwen-Video-Edit
& \textbf{1.000}
& \textbf{1.000}
& \underline{0.333}
& \underline{0.627} \\

\rowcolor{gray!10}
OneFixer
& 0.918
& 0.956
& \textbf{0.590}
& \textbf{0.219} \\

\bottomrule
\end{tabular}
\end{table}

\section{Scaling to Multi-Camera and High-Resolution Refinement}
\label{sec:challenging_generation}

\begin{figure*}[t]
    \centering
    \includegraphics[width=\textwidth]
    {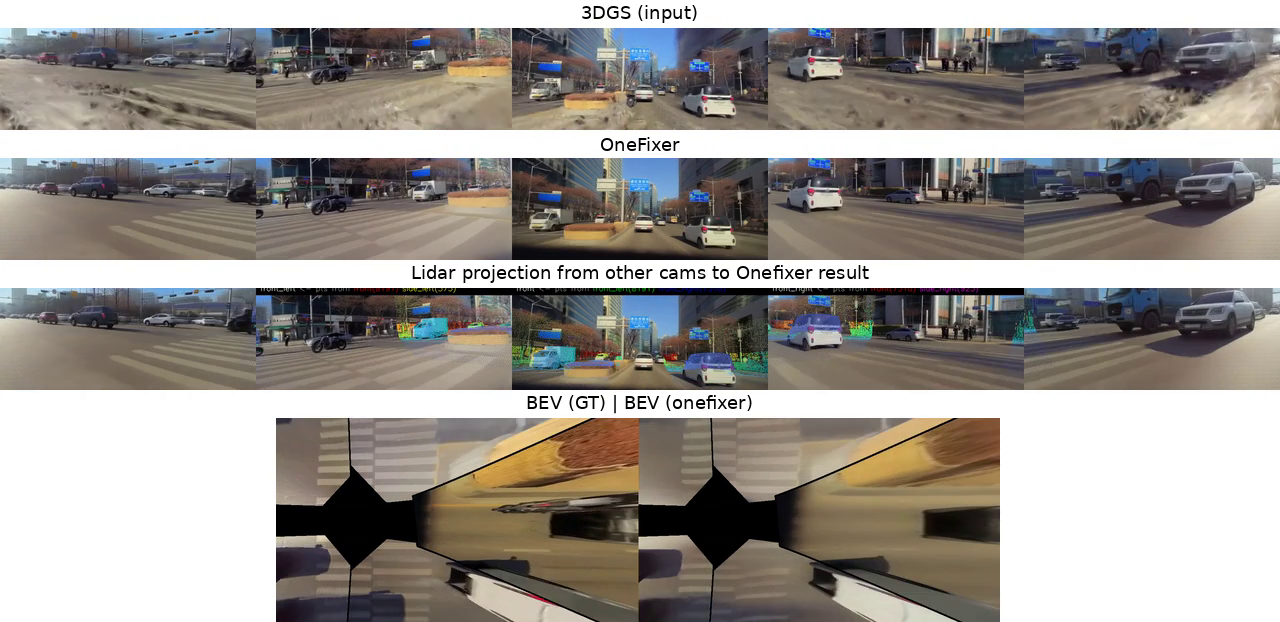}
    \caption{
    \textbf{Multi-camera 3DGS refinement.}
    From top to bottom: degraded multi-camera 3DGS renderings, OneFixer
    outputs, LiDAR points from neighboring cameras projected onto the refined
    views, and bird's-eye-view compositions from the ground truth and
    OneFixer outputs.
    OneFixer jointly removes reconstruction artifacts across views while
    preserving cross-view scene structure.
    The agreement of projected LiDAR points with the refined images and the
    resulting BEV composition provides qualitative evidence that refinement
    remains consistent with the shared underlying scene geometry.
    }
    \label{fig:multicam}
\end{figure*}

\paragraph{Multi-camera refinement.}
We extend OneFixer to synchronized multi-camera generation by adopting the
multi-view formulation of OmniDreams~\citep{nvidia2026nvidiaomnidreams}.
Joint processing across cameras substantially increases the latent sequence
length and memory footprint.
To make training feasible, we reduce the latent sequence length used per
training iteration by half and compensate with a longer optimization schedule.
Despite this reduced training horizon, the resulting model can jointly refine
multiple views while preserving cross-view scene structure.
As shown in Fig.~\ref{fig:multicam}, OneFixer removes prominent reconstruction
artifacts across views, while LiDAR points projected from neighboring cameras
remain well aligned with the refined images.
The resulting views also produce a coherent bird's-eye-view composition,
suggesting that refinement does not simply improve each camera independently
but preserves the shared scene geometry across viewpoints.

\begin{figure*}[t]
    \centering
    \includegraphics[width=\textwidth]{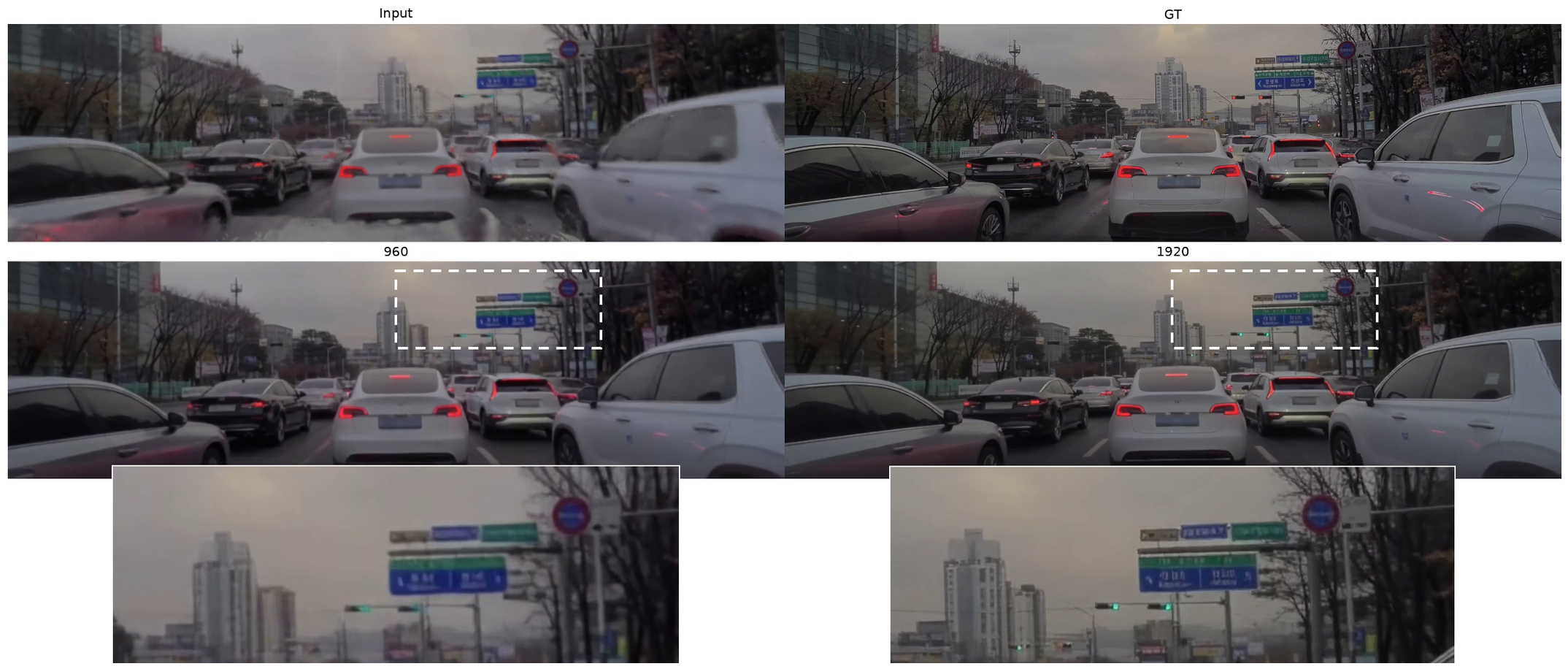}
    \caption{
    \textbf{High-resolution 3DGS refinement.}
    Top: degraded 3DGS input and clean reference.
    Bottom: OneFixer outputs at 960- and 1920-pixel widths, with enlarged
    regions highlighting distant scene details.
    Scaling to higher resolution preserves the overall refinement behavior
    while improving the recovery of fine structures such as road signs and
    other small visual details.
    High-resolution training is enabled by shortening the self-rollout segment
    during training to reduce memory consumption.
    }
    \label{fig:high_resolution}
\end{figure*}

\paragraph{High-resolution refinement.}
We further study whether the same one-step refinement formulation can scale
beyond the resolution used in our main experiments.
High-resolution training places substantially greater memory pressure on both
the autoregressive rollout and the perceptual branch.
We therefore use the same memory-efficient strategy of shortening the
self-rollout segment during training, allowing OneFixer to operate at up to
1920-pixel width.
Fig.~\ref{fig:high_resolution} compares refinement at 960- and
1920-pixel widths.
The higher-resolution model retains the overall restoration behavior of
OneFixer while recovering finer scene details, such as distant road signs and
small structural elements.
These experiments indicate that the proposed refinement framework can be
extended to more demanding spatial and multi-view settings, although doing so
requires a shorter rollout during training and therefore reduces the amount
of temporal context that can be processed in parallel.

\end{document}